\documentclass{article}
\usepackage{xcolor}

\usepackage[nonatbib,preprint]{neurips_2025}

\usepackage[utf8]{inputenc} % allow utf-8 input
\usepackage[T1]{fontenc}    % use 8-bit T1 fonts
\usepackage{hyperref}       % hyperlinks
\usepackage{url}            % simple URL typesetting
\usepackage{booktabs}       % professional-quality tables
\usepackage{amsfonts}       % blackboard math symbols
\usepackage{nicefrac}       % compact symbols for 1/2, etc.
\usepackage{microtype}      % microtypography
\usepackage{xcolor}         % colors
\usepackage{amsthm}
\usepackage{amsmath}
\usepackage{graphicx}
\usepackage{multirow}
\usepackage{amssymb}
\usepackage[official]{eurosym}
\usepackage{wrapfig}
\usepackage{algorithm}
\usepackage{algpseudocode}
\usepackage{mathtools}

\def\R{\mathbb{R}}

\def\softmax{\mathrm{softmax}}

\def\avgpoolch{\mathrm{avg\_pool\_over\_channels}}

\def\relu{\mathrm{ReLU}}

\usepackage{hyperref}
\def\our{EPIC}
\def\module{Disentanglement Module}
\def\purity{{\rm purity}}

\theoremstyle{plain}
\newtheorem{theorem}{Theorem}[section]

\theoremstyle{definition}

\theoremstyle{remark}
\newtheorem{remark}[theorem]{Remark}

\usepackage{array, makecell} %
\usepackage{arydshln}

\usepackage{subcaption}

\makeatletter
\def\adl@drawiv#1#2#3{%
        \hskip.5\tabcolsep
        \xleaders#3{#2.5\@tempdimb #1{1}#2.5\@tempdimb}%
                #2\z@ plus1fil minus1fil\relax
        \hskip.5\tabcolsep}
\newcommand{\cdashlinelr}[1]{%
  \noalign{\vskip\aboverulesep
           \global\let\@dashdrawstore\adl@draw
           \global\let\adl@draw\adl@drawiv}
  \cdashline{#1}
  \noalign{\global\let\adl@draw\@dashdrawstore
           \vskip\belowrulesep}}
\makeatother

\title{
\our{}: Explanation of Pretrained Image Classification Networks via Prototypes
}

\author{%
  Piotr Borycki \
  \thanks{ \texttt{piotr.borycki@student.uj.edu.pl}}  
    \\
    Jagiellonian University\\
  \And
  Magdalena Trędowicz\\
  Jagiellonian University\\
  \And
  Szymon Janusz\\
  Jagiellonian University\\
  \And
  Jacek Tabor\\
  Jagiellonian University\\
  \And
  Przemysław Spurek
  \\
    Jagiellonian University\\ 
    IDEAS\\
  \And
  Arkadiusz Lewicki
  \\
   University of Information Technology \\
and Management in Rzeszów\\
\And
Łukasz Struski \\
  Jagiellonian University\\
}

\begin{document}

\maketitle

\begin{abstract} 
Explainable AI (XAI) methods generally fall into two categories. Post-hoc approaches generate explanations for pre-trained models and are compatible with various neural network architectures. These methods often use feature importance visualizations, such as saliency maps, to indicate which input regions influenced the model’s prediction. Unfortunately, they typically offer a coarse understanding of the model’s decision-making process.
In contrast, ante-hoc (inherently explainable) methods rely on specially designed model architectures trained from scratch. A notable subclass of these methods provides explanations through prototypes, representative patches extracted from the training data. However, prototype-based approaches have limitations: they require dedicated architectures, involve specialized training procedures, and perform well only on specific datasets.
In this work, we propose \our{} (Explanation of Pretrained Image Classification), a novel approach that bridges the gap between these two paradigms. Like post-hoc methods, \our{} operates on pre-trained models without architectural modifications. Simultaneously, it delivers intuitive, prototype-based explanations inspired by ante-hoc techniques. To the best of our knowledge, \our{} is the first post-hoc method capable of fully replicating the core explanatory power of inherently interpretable models.
We evaluate \our{} on benchmark datasets commonly used in prototype-based explanations, such as CUB-200-2011 and Stanford Cars, alongside large-scale datasets like ImageNet, typically employed by post-hoc methods.
\our{} uses prototypes to explain model decisions, providing a flexible and easy-to-understand tool for creating clear, high-quality explanations.
% By leveraging prototypes to elucidate decisions of existing models, \our{} offers a flexible, general-purpose tool for generating high-quality, interpretable explanations.
\end{abstract}

%%%%%%%%%%%%%%%%%%%%%%%%%%%%%%%%%%%%%%%%%%%%%%%%%%%
%%%%%%%%%%%%%%%%%%%%%%%%%%%%%%%%%%%%%%%%
%%%%%%%%%%%%%%%%%%%%%%%%%%%%%
%%%%%%%%%%%%%%%%%%
%%%%%%%%
\section{Introduction}
\label{sec:introduction}

%%%%%%%%%%%%%%%%%%%%%%%%%%%%%%%%%%%%%%%%%%%%%%%%%%%
%%%%%%%%%%%%%%%%%%%%%%%%%%%%%%%%%%%%%%%%

Deep neural networks (DNNs) have revolutionized predictive modeling, frequently achieving performance superior to human experts in numerous fields \cite{he2016deep}. However, despite their impressive results, DNNs are frequently regarded as “black boxes” due to their lack of clear interpretability~\cite{lipton2018mythos}.
This lack of transparency has led to the fast development of explainable AI (XAI) methods, which aim to make accurate predictions easier for people to understand~\cite{xu2019explainable}.

\begin{figure}[tbh]
    \centering
    % \begin{minipage}[t]{0.48\linewidth}
        \centering
        \qquad\qquad\qquad \our{} (our) \qquad\qquad\qquad\qquad\quad Grad-CAM \qquad\quad LRP \\
	\includegraphics[width=0.9\linewidth]{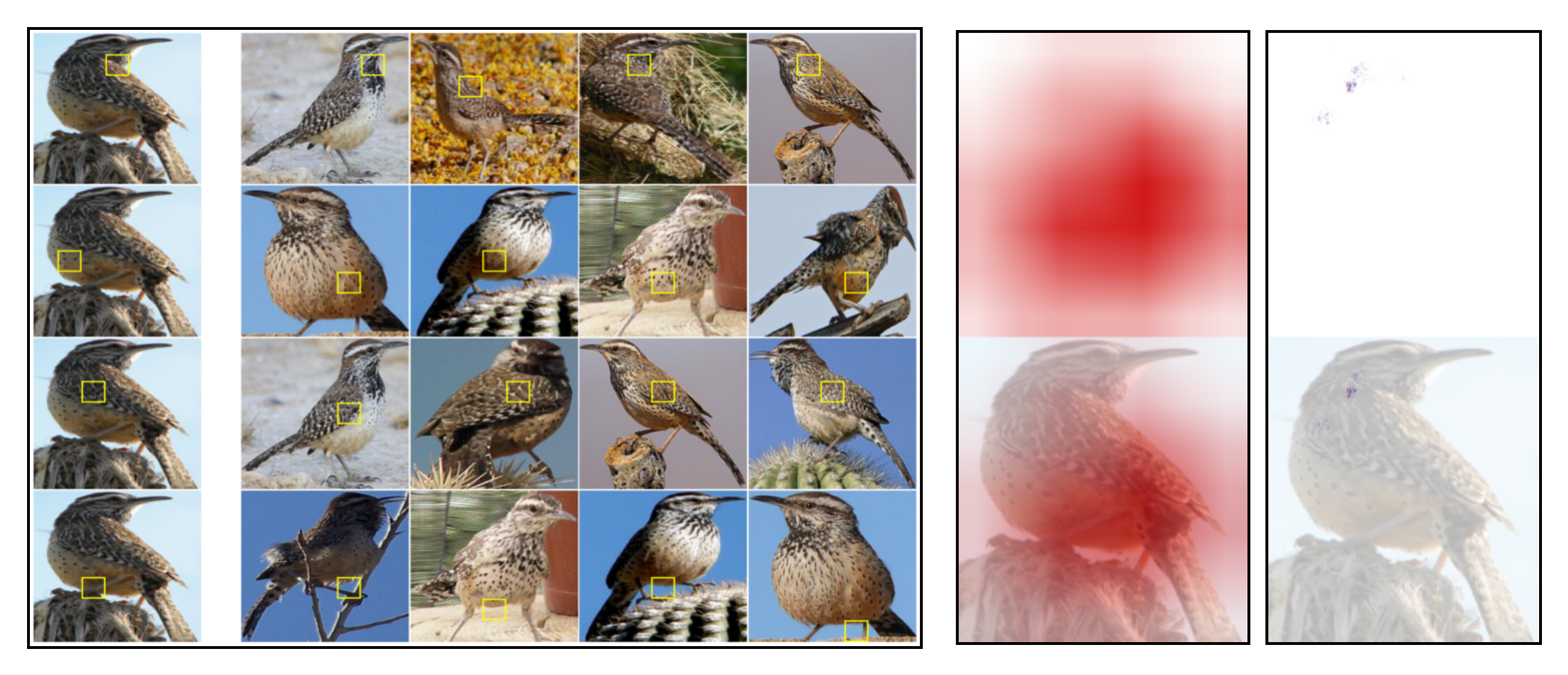}\\
        % \qquad\qquad\qquad \our{} (our) 
        % \qquad\qquad\qquad\qquad\qquad\qquad Grad-CAM \qquad\qquad LRP \\
        % \includegraphics[width=\linewidth]{NeurIPS_2025/images/epic_gradcam_lrp/epic_gradcam_lrp_imagenet.png}
    % \end{minipage}%
    % \hfill
    % \begin{minipage}[t]{0.48\linewidth}
    %     \centering
    %     \quad\qquad \our{} (our) \qquad \quad Grad-CAM \quad LRP \\
    %     \includegraphics[width=\linewidth]{NeurIPS_2025/images/epic_gradcam_lrp/epic_gradcam_lrp_imagenet.png}
    % \end{minipage}
    \caption{Comparison of explanations constructed by \our{}, and classical post-hoc models: Grad-CAM and LRP. 
    The experiment is presented in the ResNet50 feature space on the Cactus Wren image from the CUB200-2011 dataset.
    Each row of \our{} (our) represents the prototypical part. The yellow boxes in each row show the activation of a given prototypical part, while in the first column, we show the activation of corresponding prototypical parts in the original image.
    Observe that contrary to the classical XAI post-hoc approaches (Grad-CAM and LRP), \our{} provides an explanation behind the decision of the model.    
    }
    \label{fig:teaser}
    \vspace{-0.5cm}
\end{figure}

Broadly, XAI methods fall into two categories: post-hoc approaches and ante-hoc (inherently interpretable) models. Post-hoc methods apply explanation techniques to pre-trained architectures without altering their internal mechanisms. Widely adopted examples include SHAP \cite{lundberg2017unified}, LIME \cite{ribeiro2016should}, LRP \cite{bach2015pixel}, and Grad-CAM \cite{selvaraju2020grad}, all of which rely on various notions of feature importance, often visualized through saliency maps. However, while saliency maps highlight input regions contributing to predictions, they frequently fall short in providing causal or concept-level insights. As a result, they may confirm where the model is looking, but not why it arrives at a particular decision, see Fig.~\ref{fig:teaser}.

In contrast, ante-hoc (inherently explainable) models embed interpretability directly into their architectures, producing explanations as part of the prediction process. ProtoPNet \cite{chen2019looks}, a seminal example, introduced class-specific prototypes that enable explanations by comparing input image patches with prototypical parts drawn from the training data. Building on this idea, PIPNet \cite{nauta2023pip} introduced architectural and training innovations to explicitly disentangle feature channels, ensuring that each channel consistently encodes a distinct prototype. More recently, InfoDisent \cite{struski2024infodisent} leveraged a pre-trained backbone but disentangled the final layer through a modified classification head, enabling interpretable outputs without retraining the entire model.
Although ante-hoc models offer significant advantages, they encounter two fundamental challenges. First, they typically require specialized architectures and custom training regimes, demanding substantial engineering effort and computational resources. Second, they cannot be added to models that are already in use, especially if the original training data is unavailable or the model’s design cannot be changed.

\begin{figure}
    \centering
    \includegraphics[width=.85\linewidth]{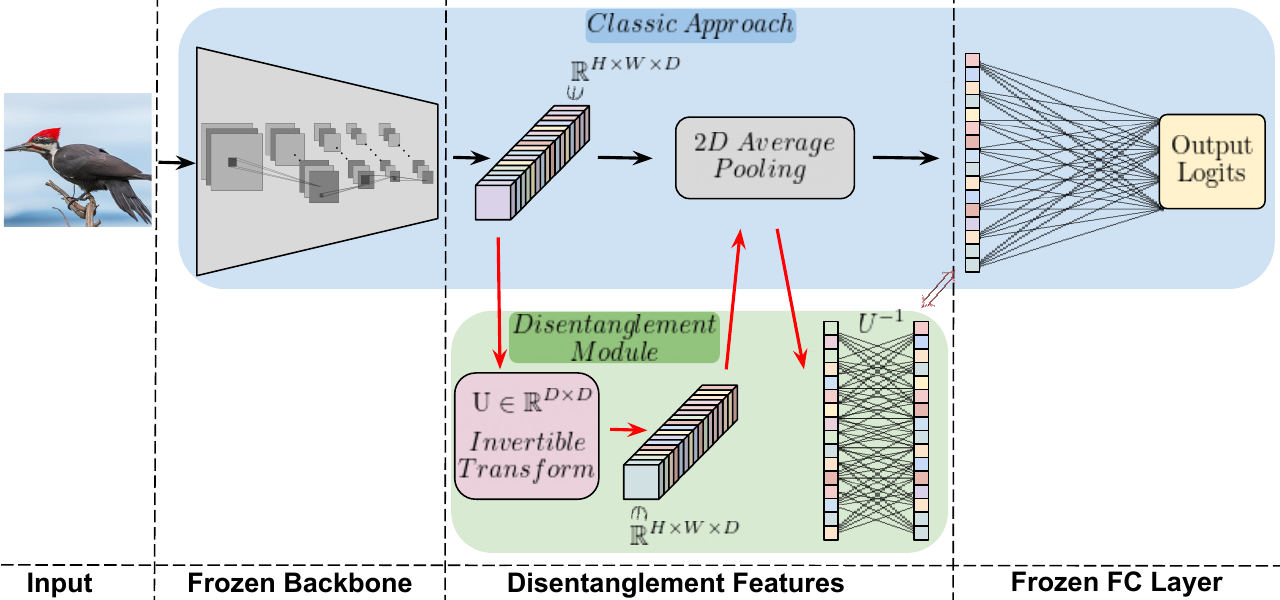}
    \caption{Our image classification interpretation model, \our{}, features three main components: a pre-trained backbone, a disentanglement layer for key features, and a fully connected layer. In contrast to the classical model, we introduce a square matrix of size equal to the number of channels, which enables disentanglement of key features. To ensure the logits remain comparable to those of the classical model, we modify the weights in the fully connected layer by multiplying them with the inverse transformation used in the feature disentanglement step.}
    \label{fig:architecture}
    \vspace{-0.3cm}
\end{figure}

In this work, we introduce Explanation of Pretrained Image Classification (\our{})\footnote{\url{https://github.com/piotr310100/EPIC}}, the first method that uses prototype-based reasoning without needing to retrain or change the original model’s design. Our approach maintains the model’s original accuracy while providing more precise and detailed explanations than typical saliency methods. We add a plugin to the model’s last layer that separates feature channels, as shown in Fig.~\ref{fig:architecture}.
\our{} is the first model that uses prototypes in post-hoc XAI models, see Fig.~\ref{fig:teaser}. Therefore, \our{} approach can be seamlessly applied to widely used datasets in prototype learning, such as CUB-200-2011 and Stanford Cars, as well as general benchmarks like ImageNet, demonstrating broad applicability across tasks.

\begin{figure}[!t]
    \centering
    \qquad Before \our{} optimization \qquad\qquad\qquad\qquad\qquad After \our{} optimization \\
        \includegraphics[width=0.45\linewidth]{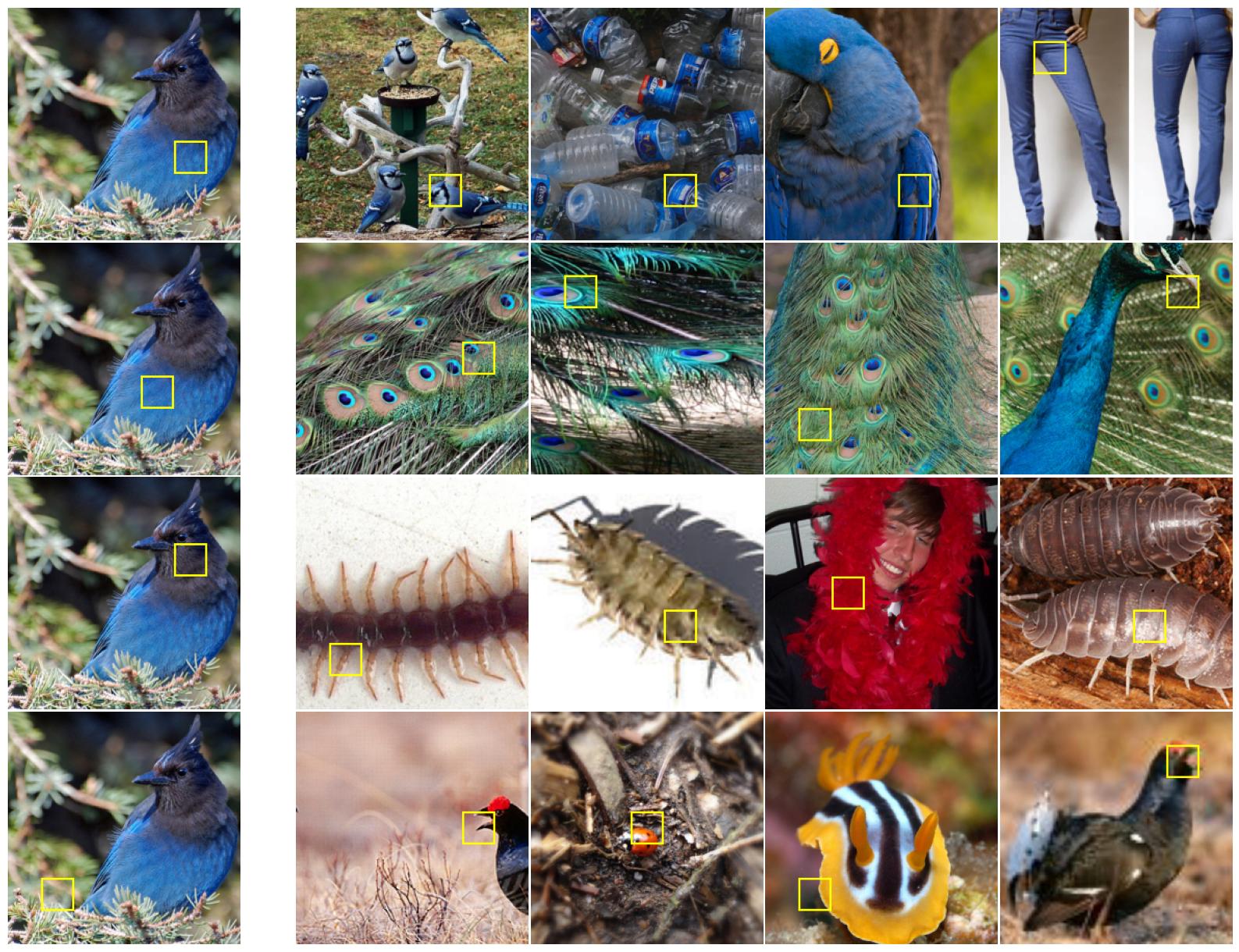}
        \qquad
        \includegraphics[width=0.45\linewidth]{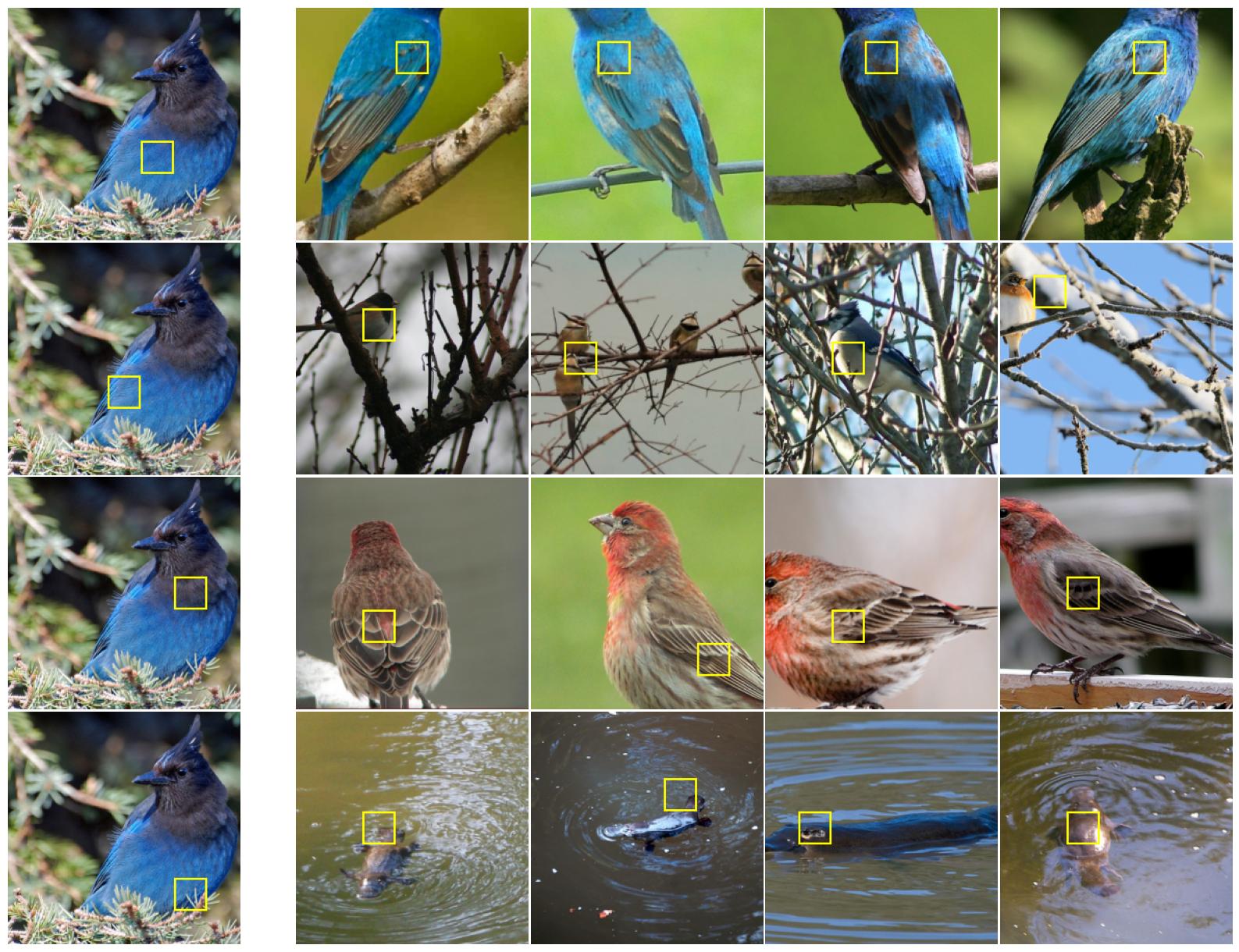}
    \caption{Explanations for a blue jay bird, before (left) and after (right) \our{} training on Resnet18.
    As we can see, prototypes without additional tuning correspond to random images and have limited explanatory properties. After \our{} tuning, such prototypes are consistent and correspond with input image features. 
    % \piotrek{TODO: odwołanie w tekście}
    }
    \vspace{-0.3cm}
    \label{fig:expl_before_after_imagenet}
\end{figure}

The core idea behind \our{} centers on defining a prototype purity measure, quantifying the degree of disentanglement of feature channels in the final layer. Naively extracting prototypes from a standard trained model typically results in low-quality explanations, as the learned channels are not aligned with coherent, interpretable concepts, see Fig.~\ref{fig:expl_before_after_imagenet}. To address this, \our{} introduces a lightweight sub-module attached to the final layer, which selectively reshapes the channel representations based on purity criteria. Crucially, this enhancement operates without altering the model’s predictions, focusing solely on producing disentangled, meaningful prototype channels.
Our key contributions are summarized as follows:
\begin{itemize}
\item We propose \our{}, a principled post-hoc explanation framework that integrates prototype-based reasoning into existing deep models without retraining.
\item We demonstrate that \our{} offers superior interpretability over saliency-map-based approaches by explicitly targeting prototype purity.
\item We validate the versatility and generality of \our{} on both specialized fine-grained datasets (CUB-200-2011, Stanford Cars) and large-scale classification tasks (ImageNet).
\end{itemize}

%%%%%%%%%%%%%%%%%%%%%%%%%%%%%
%%%%%%%%%%%%%%%%%%
%%%%%%%%

\section{Related Works}
\label{sec:related_works}

With the dynamic development and increasingly widespread deployment of deep learning models in key areas such as healthcare, finance, and autonomous systems, the issue of explainability has acquired the status of a fundamental research challenge. In the scholarly literature on explainable artificial intelligence (XAI), two principal paradigms can be distinguished: post-hoc explanation methods and inherently interpretable (ante-hoc) models. 

Post-hoc methods focus on analyzing already trained models, providing explanations without interfering with their architecture. One example of such a method is SHAP (SHapley Additive exPlanations), which employs Shapley values to assign importance to individual features in a model's prediction \cite{lundberg2017unified}. Similarly, the LIME (Local Interpretable Model-agnostic Explanations) method enables the creation of local linear models to interpret predictions \cite{ribeiro2016should}. Techniques such as Grad-CAM (Gradient-weighted Class Activation Mapping) generate attention maps that highlight input regions critical to the model’s decision-making process \cite{selvaraju2017grad}. However, despite their popularity, these methods are often criticized for the instability and inconsistency of the explanations they generate, as well as for their limited ability to capture causal relationships \cite{adebayo2018sanity}.

By contrast, ante-hoc models integrate interpretability mechanisms directly into the architecture of the model itself. One such development is the ProtoPNet (Prototypical Part Network) algorithm, which introduces the concept of class prototypes, allowing the interpretation of model decisions by comparing image segments to representative prototypes \cite{chen2019looks}. Extensions of this approach, such as PIPNet (Prototype Interpretable Part Network), introduce mechanisms for prototype selection and channel decomposition, thereby improving the quality of interpretations achieved \cite{nauta2023pip}. Nevertheless, ante-hoc models often require specialized architectures and retraining, which limits their applicability in existing, complex systems.

In response to the limitations of both approaches mentioned above, hybrid methods have been proposed. These combine the advantages of post-hoc and ante-hoc techniques. In this area, recent years have seen the development of solutions such as ACE (Automated Concept-based Explanations) and Concept Whitening. The ACE algorithm automatically identifies semantically coherent concepts within network layers, providing human-understandable interpretations \cite{ghorbani2019automating}. Meanwhile, Concept Whitening introduces a mechanism for orthogonalizing the latent space, enabling a better understanding of the model’s internal representations \cite{chen2020concept}. Although these methods offer new interpretability opportunities, they often do not provide prototype-based explanations characteristic of ante-hoc approaches and acceptable as fully correct.

Thus, there exists a clear gap between the flexibility of post-hoc methods and the deep interpretability of ante-hoc models. Our proposed method addresses this gap by enabling prototype-based explanations on top of already trained models. It combines the scalability offered by post-hoc techniques with the interpretability characteristic of ante-hoc approaches. Importantly, it achieves this without requiring any architectural modifications or retraining.

\section{\our{}: Explanation of Pretrained Image Classification}
\label{sec:method}

In this section, we present the \our{} model, designed specifically to provide explanations for deep neural networks. Our approach involves integration of a plug-in \module{} into the network’s final layer, the classification head. \our{} disentangles the feature channels in this last layer based on a purity measure. As a post-hoc method, our model is applied to explain neural networks that have already been trained.

% In this section, we describe \our{} model, which is dedicated to explaining deep neural networks. In practice, our model adds \module{}, which is plugged into the last layer -- classification head. \our{} disentangle feature channels in the final layer according to the purity measure. Our model is a post-hoc model, and explains previously trained models. 

Our paper considers the classification networks used in PIPNet \cite{nauta2023pip} and InfoDisent \cite{struski2024infodisent}. In the case of a classification task with $k$ classes, we assume that we have a backbone $\Phi_{\Theta}$ that transforms the input image $I$ into the feature space $\Phi_{\Theta}(I)\in \R^{H\times W\times D}$ where $H,W$ denote height and width of the map, and $D$ denotes the number of channels (depth). Such a feature map then undergoes the pooling operation 
$$
v_I=\avgpoolch(\Phi_{\Theta}(I)) \in \R^{D}.
$$
At the end of such operations, we have a linear classification layer $w_I=Av_I$, where $A$ is a matrix of dimensions $N \times D$, where $N$ is the number of classes. Finally, Softmax is applied to obtain the final probabilities for each class.

In this type of architecture, each channel of the final feature space in which the $\Phi_{\Theta}(I)$ resides can be interpreted as an individual prototype \cite{nauta2023pip,struski2024infodisent}. Before explaining how to ensure these channels provide coherent explanations, we first demonstrate the process of finding prototypes of a fixed channel for a traditionally trained model. Subsequently, we introduce a measure for the distribution of the channels in a prototype, referred to as the purity measure. We then describe the approach to maximize the purity using \module{}. Finally, we outline the construction of the explanations for an input image. 

\begin{figure}
    \centering
    \includegraphics[width=\linewidth]{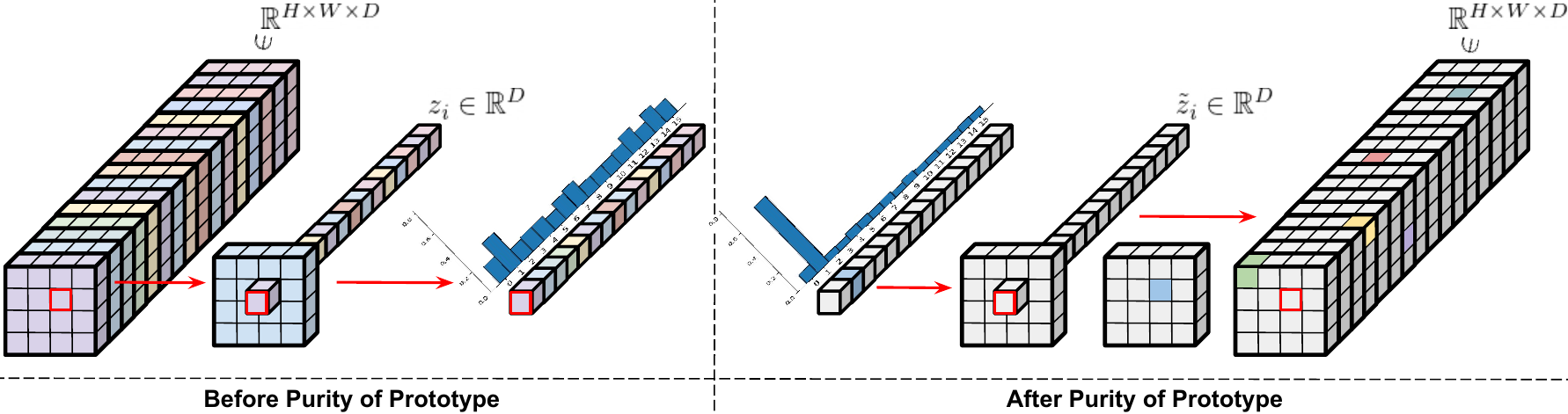}
    \caption{
        The illustration demonstrates the concept of the \emph{Purity of Prototype} mechanism. 
        For a selected channel, the vector $\mathbf{z}$ (shown on the left) is defined by the maximum pixel value in that channel, making its values \emph{comparable} (histogram of activation is flat). 
        After optimizing the purity of the given prototype, only one dominant value remains in the refined vector $\mathbf{\tilde{z}}$, as seen on the right. 
        Repeating this process for each channel results in a disentangled representation, where each channel contains only one dominant value associated with its prototype.
    }
    \label{fig:purity_prototype}
\end{figure}

% \begin{figure}[tbh]
%     \centering
%     % \begin{minipage}[t]{0.32\linewidth}
%         % \centering
%         % \includegraphics[width=0.45\linewidth]{NeurIPS_2025/images/epic_gradcam_lrp/dogs/epic_gradcam_lrp_Airedale_4_framed.png}
%     % \end{minipage}
%     % \begin{minipage}[t]{0.32\linewidth}
%         % \centering
%         % \qquad
%         \qquad\qquad\qquad \our{} (our) \qquad\qquad\qquad\qquad\quad Grad-CAM \qquad\quad LRP \\
%         \includegraphics[width=0.9\linewidth]{NeurIPS_2025/images/epic_gradcam_lrp/imagenet/epic_gradcam_lrp_hognose snake_4_framed.png}
%     % \end{minipage}
%     % \begin{minipage}[t]{0.32\linewidth}
%     %     \centering
%     %     \includegraphics[width=\linewidth]{NeurIPS_2025/images/explanations/cub/resnet50/after/explanations_Pine Warbler.png}
%     % \end{minipage}
%     \caption{Explanations for the Hognose Snake from ImageNet  constructed by \our{}(our), Grad-CAM and LRP. \our{} prototypes effectively capture crucial concepts, such as shapes, colors, textures and distinctive features like snake's eyes area. In contrast, existing post-hoc methods (Grad-CAM and LRP) produce only saliency maps, offering less interpretability regarding specific visual attributes and concepts.}
%     \label{fig:expl_snake}
% \end{figure}

\paragraph{Prototypes of a feature map channel}
% \label{subsec:prot_channel}
The main component of our approach is finding a set of images connected to each feature map channel, which will represent the information propagated by a specific channel. Consequently, we are looking for $m$ (usually $m=5$) prototype images from the training set for a fixed channel $k$. All that remains is to specify how the prototypes are selected. Provided an image $I$ we calculate its representation in the feature space 
$
Z_I = \Phi_{\Theta}(I) \in \R^{H\times W\times D}.
$
% Such representation can be seen as a feature map with $D$ channels of dimension $W\times D$.
This can be viewed as a representation on which the model's classification head works.

We are looking for $m$ images that activate mainly on the $k$-th channel. More specifically, we define the activation of a channel $k\in\{1,2,\ldots,D\}$: 
$$
    \operatorname{activ}(Z; k)= \sum_{h=1}^{H}\sum_{w=1}^{W} Z[h,w,k] \; \mbox{ for feature map } Z\in\R^{H\times W\times D}.
$$
Activation of the channel $k$ at height $h$ and width $w$ in the feature space is denoted by $Z[h,w,k]$. Additionally, let us note that we will later refer to the vector $Z[h,w]\in\R^D$ as a pixel in feature space interpreted as an image with $D$ channels. This vector will later be crucial to understanding the prototype's quality.

By using channel activation, we can select $m$ prototype images for the $k$-th channel:
    \[
    \text{Prot}_{\text{pos}}^{(k)} = \operatorname{arg\,top-m}_{I \in \text{TrainSet}} \operatorname{activ}(Z_I; k).
    \] 
This process can be summarized as the application of the channel activation function to all images in the training set, and finding the images for which the $m$ largest values is obtained. The chosen images will be called positive prototypes of channel $k$. Similarly we can define negative prototypes as 
    \[
    \text{Prot}_{\text{neg}}^{(k)} = \operatorname{arg\,top-m}_{I \in \text{TrainSet}} -\operatorname{activ}(Z_I; k).
    \]

This process can be repeated for all channels to obtain their prototypes. The results for the classically trained neural network without any modifications and the results of \our{} are presented in Fig. \ref{fig:expl_before_after_imagenet}. As we can see, without additional tuning, such prototypes are less clear than the ones obtained after the training of \our{}. To measure the quality of the prototype image we use a measure called purity introduced in the following section. In our model, we use \module{} to make the prototypes more coherent. However, we still have to find a method to evaluate the quality of a prototype.

\paragraph{Purity of prototype}
In this paragraph, we define the purity measure employed by \our{} to disentangle channels in the feature space. Classical optimization concentrates on the prediction task and produces a mixed representation. As a result, concepts related to the model prediction are entangled between different channels. Representation is fully disentangled if only one channel is active for a given image. 
\our{} uses purity measure to assess the disentanglement of the future space, see Fig.~\ref{fig:purity_prototype}. In our paper, we focus on the positive prototypes. However, the process is analogous for negative prototypes. Below, we present a detailed formulation of the purity of the prototype.

For a given backbone $\Phi_{\Theta}$, input image $I$, and 
selected prototypical channel $k$, we define a prototypical pixel, the coordinates of it are defined as
\[
    \mathbb{N}^2 \ni(h,w) = \arg\max_{x,y} Z_I[x,y,k].
% M_I[h_{max}(i), w_{max}(i), \cdot] \in \R^d. 
\]
That is the coordinates of the largest activation in the $k$-th channel. The prototypical pixel is then given by a vector
$
    p = Z_I[h,w]\in\mathbb{R}^D.
$
It spans the channels across the spatial location in which the largest activation of $k$-th channel is achieved. By using this vector we can define the purity by:
\[
\purity\left(I,k\right) = \frac{p_k}{ \| p \| }\in[0,1].
\]
If the value of $\purity\left(I,k\right)$ is equal to one, we call the prototype pure. This situation occurs, when all but the $k$-th channel activations are zeroes, which is consistent with the motivation behind this measure. In Fig.~\ref{fig:purity_prototype}, we visualize such a situation. Before purity optimization, our prototype pixels were not pure since the histogram of activation was uniformly distributed. After optimization, the neural network activates mainly on a single coordinate. During optimization of \module{} the feature space is disentangled by forcing the prototypes to be pure.

\begin{figure}[!t]
    \centering
    % \begin{minipage}[t]{0.32\linewidth}
        % \centering
        % \includegraphics[width=0.45\linewidth]{NeurIPS_2025/images/epic_gradcam_lrp/dogs/epic_gradcam_lrp_Airedale_4_framed.png}
    % \end{minipage}
    % \begin{minipage}[t]{0.32\linewidth}
        % \centering
        % \qquad
        \qquad\qquad\qquad \our{} (our) \qquad\qquad\qquad\qquad\quad Grad-CAM \qquad\quad LRP \\
        \includegraphics[width=0.9\linewidth]{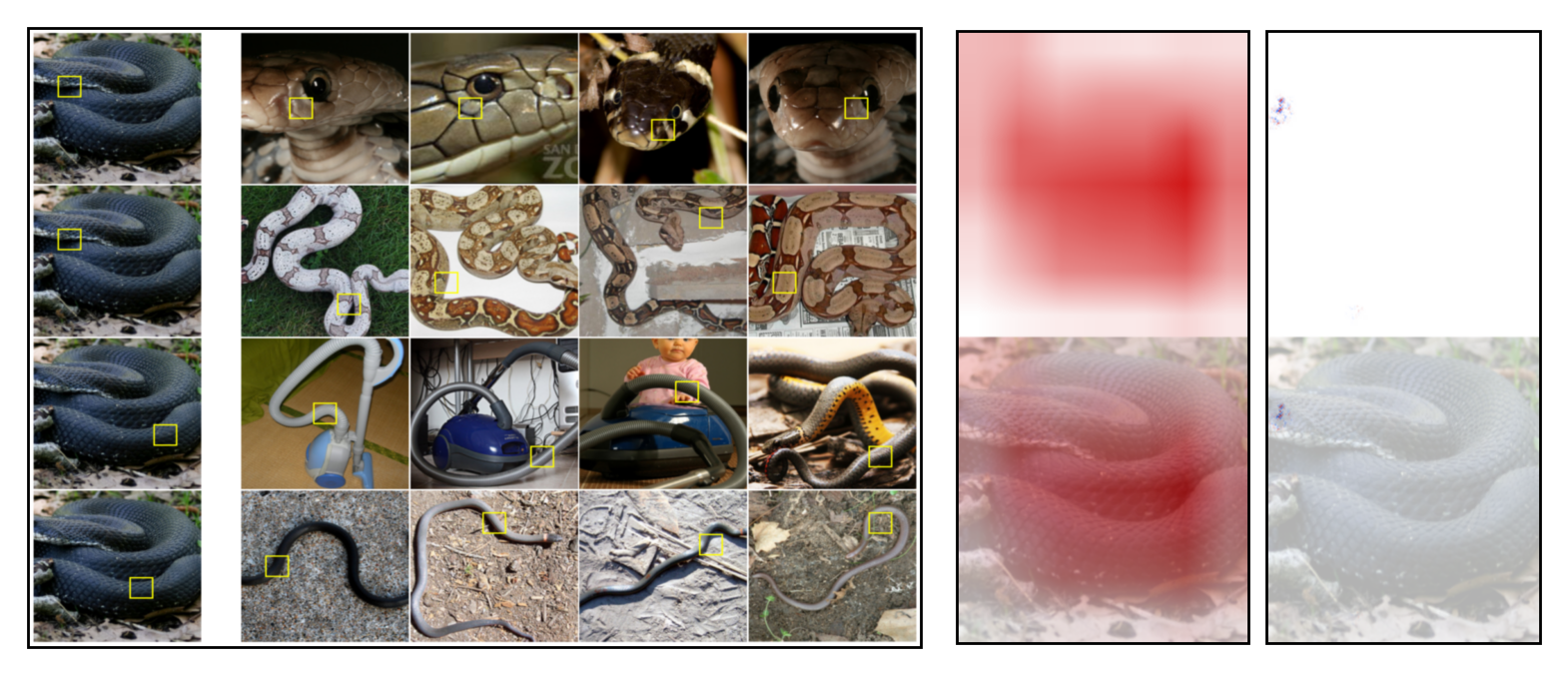}
    % \end{minipage}
    % \begin{minipage}[t]{0.32\linewidth}
    %     \centering
    %     \includegraphics[width=\linewidth]{NeurIPS_2025/images/explanations/cub/resnet50/after/explanations_Pine Warbler.png}
    % \end{minipage}
    \caption{Explanations for the Hognose Snake from ImageNet  constructed by \our{} (our), Grad-CAM and LRP. \our{} effectively capture crucial concepts, such as shapes, colors, textures, and distinctive features like the snake's eye area. In contrast,  Grad-CAM and LRP produce only saliency maps, offering less interpretability regarding specific visual attributes and concepts.}
    \label{fig:expl_snake}
    \vspace{-0.3cm}
\end{figure}

% \begin{figure}
%     \centering
%     \includegraphics[width=\linewidth]{NeurIPS_2025//images/purity.pdf}
%     \caption{
%         The illustration demonstrates the concept of the \emph{Purity of Prototype} mechanism. 
%         For a selected channel, the vector $\mathbf{z}$ (shown on the left) is defined by the maximum pixel value in that channel, making its values \emph{comparable} (histogram of activation is flat). 
%         After optimizing the purity of the given prototype, only one dominant value remains in the refined vector $\mathbf{\tilde{z}}$, as seen on the right. 
%         Repeating this process for each channel results in a disentangled representation, where each channel contains only one dominant value associated with its prototype.
%     }
%     \label{fig:purity_prototype}
% \end{figure}

\paragraph{\module{}}
The prototypes can be used to explain a neural network's prediction, as noted the larger the purity the better the explanation. Our goal is to disentangle channels in the feature space of a pretrained model, while simultaneously preserving the original models prediction. Consequently we propose to use a \module{}, which uses a learnable invertible matrix $U\in\mathbb{R}^{D\times D}$ to separate the channels inside the feature space. Thus, \our{} is injected into the model just before the Pooling Layer, and the final linear layer weight is multiplied by $U^{-1}$ to preserve the original output. More precisely, for an input image $I$, we first transform the original image into feature space $Z=\Phi_\Theta(I)\in\mathbb{R}^{H\times W\times D}$. Next, we apply the matrix $U\in\mathbb{R}^{D\times D}$ to each spatial location of $Z\in\mathbb{R}^{H\times W\times D}$, transforming feature space in which the channels are disentangled. More precisely, for each pixel coordinates $(x,y)$ the feature vector $Z[x,y]\in\mathbb{R}^D$ is projected to a new space by
$
    \mathbb{R}^D\ni\hat{Z}[x,y]=U Z[x,y].
$
This operation can be summarized as the application of a linear operator $U$ to each pixel. We will later denote this operation by $U \circledast Z$.

% This operation changes the models output. 
To preserve the original activations, we have to reverse this operation in the classification head of the model. This can be achieved by substituting the weight $A$ of the linear classification layer, by $A'=AU^{-1}$. The final model can be summarized as
\begin{align}
    & Z=\Phi_\Theta(I)\in\mathbb{R}^{H\times W\times D}, \\
    & \hat{Z}= U \circledast Z \in\mathbb{R}^{H\times W\times D} ,\; U\in\mathbb{R}^{D\times D},\\
    & v=\avgpoolch(\hat{Z}) \in \R^{D}, \\ 
    & w = A'v=(AU^{-1})v, \\
    & \operatorname{pred} = \softmax (w).
\end{align}

The above neural network modification does not change the final prediction of the network, which is a consequence of the simple Remark~\ref{remark:1}.

\begin{remark}
\label{remark:1}
Let $Z\in\mathbb{R}^{H\times W\times D}$ be an image representation in feature space and $U\in\mathbb{R}^{D\times D}$ an invertible matrix, than:
$$
U^{-1}\mathrm{avg\_pool\_over\_channels}(U \circledast Z) = \mathrm{avg\_pool\_over\_channels}(Z).
$$
\end{remark}
\begin{proof}
This follows from a distributative property of matrices. At each spatial location $(x, y)$, we have:
$$
\begin{array}{c}
    U^{-1} \avgpoolch(U \circledast Z)
    = U^{-1} \left( \dfrac{1}{HW} \sum_{x,y} U Z(x,y) \right) =\\ 
    = U^{-1} U \left( \dfrac{1}{HW} \sum_{x,y} Z(x,y) \right) 
    = \avgpoolch(Z).
    \\
    \vspace{-0.5cm}
    \qedhere
\end{array}
$$ 
% \vspace{-0.5cm}
\end{proof}

Such a simple modification allows us to disentangle channels. We train the matrix $U$ with a restriction to either the class of invertible or orthogonal matrices. It is worth noting that if we set the matrix $U$ to identity matrix, we get exactly the original pretrained model. 

\paragraph{Training}

As mentioned in the previous section the quality of a prototype is tied to the value of purity. Consequently, the training stage focuses on the maximization of prototypes purity. But since, we want to preserve the original model output, all its weights are frozen, and only the elements of matrix $U$ in the \module{} are updated. Additionally, the optimization process is done solely on the set of prototypes. However, since each update to matrix $U$ causes a change in the activations of channels, the new set of prototypes has to be recalculated every few epochs throughout the training. This provides the compromise between the speed, and dynamic updates to prototypes. In our experiments, the \module{} was trained for 20 epochs, with prototypes being recalculated every 2 epochs. In addition to the update of prototypes, the number of prototypes for each channel is decreased at the same time. We start with 100 images for each prototypical channel, and linearly decrease this value to 5 at the end of the training stage.

\paragraph{Explaining model prediction}
After completing the training of the \module{} and selecting the channel prototypes, the next step is to explain the model's prediction for a given input image. This is achieved by selecting $k$ channels with the highest contribution to the predicted class. This can be done by examining the terms contributing to the model output in the final classification layer. More precisely, for an input image $I$ and the model prediction of the input belonging to class $y$ (for more details, see the algorithm in the Appendix~\ref{app:model_pred}). %algorithm \ref{alg:explain}. 
Since we are only interested in the positive prototypes, we apply $\relu$ before examining the terms contributing to the sum. Example explanation is shown in Fig. \ref{fig:expl_snake}.

% During training, the only learnable parameter is the matrix $U$, and the original model weights are frozen. It is initialized as an identity matrix. The dataset used for training is the set of $n$ prototypes for each dimension in the feature space. The prototypes are chosen from  the original training split used to obtain the pretrained model. However, only the inference time transformations are applied to each image. We use multiple values of $n$ during training. Initially a large value $N$ is chosen, and during training it gradually decreases over a span of $t$ steps to a final number of prototypes for each channel $m$. Typically we used the values of $N=100$ and $m=5$, while the amount of steps $t=10$. During each step the new prototypes are found, and their union is used for training over a small number of epochs (we used at most 2 epochs in our experiments). This process ensures that the prototypes are dynamically changed, whenever a new image is a better candidate.

% \section{Method/Theory}
% \label{sec:method_theory}

% \begin{itemize}
%     \item \todo{opis metody, miara oceny}
%     \item \todo{obrazek opisujący działanie purity}
% \end{itemize}

%%%%%%%%%%%%%%%%%%%%%%%%%%%%%%%%%%%%%%%%%%%%%%%%%%%
%%%%%%%%%%%%%%%%%%%%%%%%%%%%%%%%%%%%%%%%
%%%%%%%%%%%%%%%%%%%%%%%%%%%%%
%%%%%%%%%%%%%%%%%%
%%%%%%%%
\section{Experiments and Results}
\label{sec:ex_results}

In the experimental section, we evaluate our model across several scenarios. First, we provide a qualitative comparison, showcasing example predictions and comparing our results against post-hoc methods such as Grad-CAM, LPR. We also compare our model to the prototype-based model InfoDisent, which works with the ImageNet dataset.
 Then, we present that our model is only a plugin to the model, and we do not change the network's prediction. Next, we show a multidimensional analysis of the FunnyBirds datasets. Finally, we present the results of user studies.

% \begin{wrapfigure}[11]{r}{0.5\linewidth}
%   \centering
%   \begin{minipage}{\linewidth}
%     \includegraphics[width=\linewidth]{NeurIPS_2025/images/channel_prototypes/imagenet/resnet34/before/channel_114.png}\\
%     \includegraphics[width=\linewidth]{NeurIPS_2025/images/channel_prototypes/imagenet/resnet34/after/channel_114.png}
%   \end{minipage}
%   \caption{Prototypical images before and after the training of \module{} for ResNet34}
%   \label{fig:prot_resnet34}
% \end{wrapfigure}

% \paragraph{Prototypes of feature channels}

% Feature map channels play a crucial role in understanding the model decisions using \our{}. By the prototypical images of channel $k$ we understand a set of $m$ images with the largest channel activation as explained in \ref{subsec:prot_channel}. Initially before the training of \module{} the prototypes rarely represent coherent information. This is fixed by training the matrix $U$ to maximize the purity of prototypes (see Fig. \ref{fig:prot_resnet34}).

\paragraph{Explanation of model decision}

This section outlines the experimental results of \our{} explanations and its comparison to other XAI methods, including both post-hoc and ante-hoc approaches. 
Fig.~\ref{fig:expl_snake} illustrates the interpretability improvements of \our{} over classical post-hoc methods, Grad-CAM and LRP, on the imput images from CUB200-2011 and Stanford Dogs datasets. Each row in the \our{} visualization represents the prototypical part (the corresponding channel number). The yellow boxes in each row show the activation of a given prototypical part, while in the second column, we show the activation of corresponding prototypical
parts in the original image. While \our{} demonstrates clear part-level interpretability, Grad-CAM and LRP produce more diffused heatmaps that highlight general areas of importance but lack the fine-grained interpretability provided by \our{}. Grad-CAM and LRP can identify important regions only within an input image and they fall short of capturing visually meaningful concepts such as textures, shapes, and distinctive object parts across different samples from the dataset. In contrast, \our{} not only highlights critical regions but also provides semantically rich prototypes that represent these crucial visual features. Additional examples can be found in Appendix~\ref{app:model_pred}.

Fig.\ref{fig:comp_mushroom} presents a comparison of explanations generated by \our{} and the prototype-based model InfoDisent. While InfoDisent operates on a pretrained backbone and can produce predictions on the ImageNet dataset, \our{} constructs prototypes that are more closely aligned with the input images.

% Examples of these can be easily observed in Figure~\ref{fig:expl_snake}. We can note the precision with which \our{} explains the model's decisions. For the left input image from the Stanford Dogs dataset, \our{} identifies features relevant to the input breed (Airedale), such as the texture and coloration of the fur, as well as the areas around the legs and head. The middle input image from ImageNet is explained by \our{} based on the snake's head, skin texture, and the surrounding ground. Notably, the coiled shape of the snake is highlighted, as it bears resemblance to a vacuum cleaner hose. The Pine Warbler (CUB200-2011) was correctly classified based on the feathered regions. It is important to note that this species, along with many other bird species, exhibits a variety of colors and patterns on its body, which \our{} successfully captures, including the yellow coloration of the tummy and neck, as well as the mottled pattern on the wings.

% dopisać o figure 7: infodiscent vs epic 

\begin{figure}
    \centering
    \qquad \our{} (our) \qquad\qquad\qquad\qquad\qquad\qquad InfoDisent \\
    \includegraphics[width=0.9\textwidth]{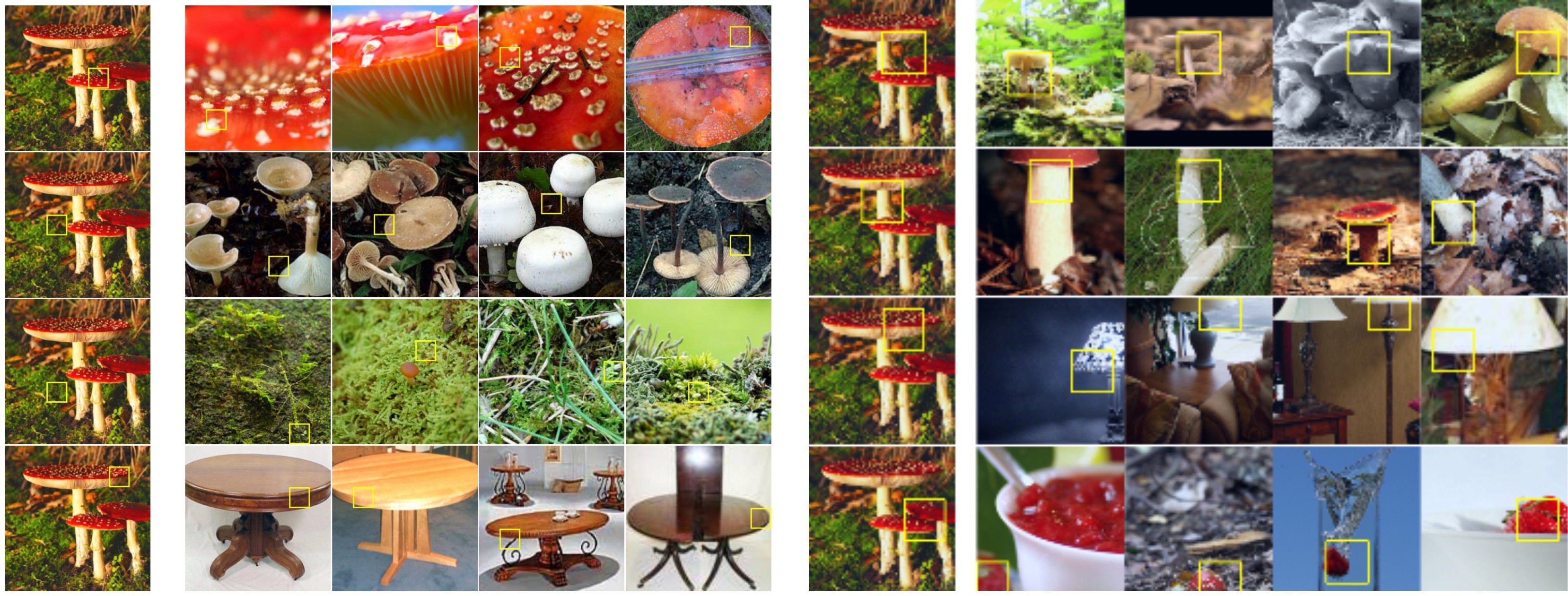}
    \caption{Comparison of explanations between \our{} (our) and prototype-based model InfoDisent. 
    InfoDisent works on top of the pretrained backbone and can give predictions for the ImageNet dataset. \our{} build prototypes more connected with input images. The comparison is conducted on a representation learned on top of pretrained ResNet50.}
    \label{fig:comp_mushroom}
    \vspace{-0.3cm}
\end{figure}

\begin{wraptable}{r}{0.5\textwidth}
  \centering
%  \vspace{-1.3cm}
  \caption{Classification accuracy (ACC) on ImageNet dataset by competing approaches using different backbones.}
  \footnotesize
  \begin{tabular}{l@{\;\,}c@{\qquad\;}l@{\;\,}c}
    \toprule
    \makecell{\bf Model} & \textbf{ACC} & \makecell{\bf Model} & \textbf{ACC} \\
    \midrule
    ResNet-34 & 73.3\% & ConvNeXt-L & 84.4\%    \\
    \our{} & \textbf{73.3}\% & 
    \our{} & \textbf{84.4} \%  \\
    InfoDisent & 64.1\% & 
    InfoDisent & 82.8\% \\
    \cdashlinelr{1-4}
    ResNet-50 & 80.8\%  & Swin-S & 83.7\%  \\
    \our{} & \textbf{80.8}\% & \our{} & \textbf{83.7}\% \\
    InfoDisent & 67.8\% & InfoDisent & 81.4\% \\
    \cdashlinelr{1-4}
    DenseNet-121 & 74.4\%  & &\\
    \our{} & \textbf{74.4} \% & 
    &  \\
    InfoDisent & 66.6\% & 
    &   \\
    % \cdashlinelr{1-4}
    % MaxVit & 83.4\% & &\\
    % \our{} & 83.3\%  & &\\
    \bottomrule
  \end{tabular}
  \vspace{-0.5cm}
  \label{tab:acc_imagenet}
\end{wraptable}

\paragraph{Classification Performance}
As previously mentioned, the construction of \our{} preserves the predictive ability of the pretrained model. This means that \module{} does not change the model output. However, since we apply additional operations, numerical errors might arise. To show that this situation does not occur, we present in Tab.~\ref{tab:acc_imagenet} the numerical accuracy on ImageNet. Results on various datasets are presented in Appendix~\ref {app:class_perform}.

\paragraph{Multi-dimensional analysis}

\begin{figure}
  \centering
    \includegraphics[width=.99\linewidth]{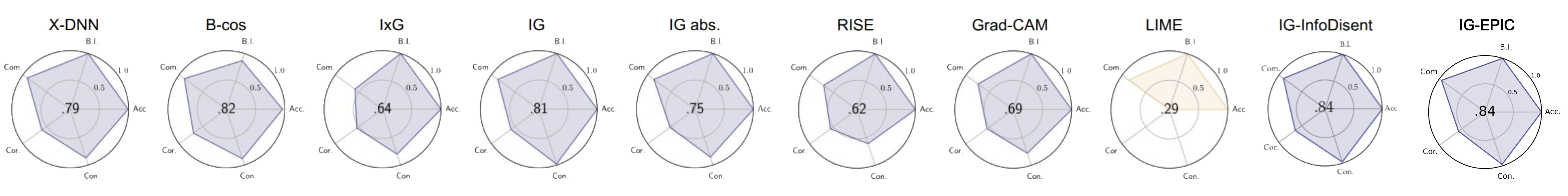}\\
  \caption{FunnyBirds evaluation results for various XAI methods: Input$\times$Gradient (IxG) \cite{shrikumar2017justblackboxlearning}, (absolute) Integrated Gradients (IG (abs.)) \cite{sundararajan2017axiomatic}, Grad-CAM \cite{selvaraju2017grad}, RISE \cite{petsiuk2018rise}, LIME \cite{ribeiro2016should}, $X$-DNN \cite{hesse2021fast}, B-cos network \cite{bohle2022b} and InfoDisent \cite{struski2024infodisent}. Resnet50 are used to evaluate model-agnostic techniques. The center score, which represents the mean of the completeness (Com.), correctness (Cor.), and contrastivity (Con.) dimensions, is calculated by averaging the results throughout the whole test set. Furthermore, background independence (B.I.) and accuracy (Acc.) are reported. Our approach (last from the left) equals the best result for Resnet50.}
  \label{fig:funny_birds}
  \vspace{-0.5cm}
\end{figure}

To assess our methodology, in the last experiment, the FunnyBirds \cite{Hesse:2023:FunnyBirds} dataset was used. Semantically relevant image modifications, like deleting individual object pieces, are supported by the FunnyBirds dataset as well as by our innovative automatic evaluation algorithms. Thus,  XAI methods and model architectures were developed to provide a more comprehensive evaluation of explanations on the part level. Like humans observing an image, they concentrate on distinct elements instead of individual pixels. \our{} is compared with multiple methods on classical convolutional network (Resnet50) for which it ranks among the best Fig.~\ref{fig:funny_birds}.

\paragraph{User study results}
We conducted two user studies, each involving 60 participants per dataset. Both studies utilized two datasets: CUB-200-2011 and ImageNet. During the studies, each participant answered 20 questions, with images randomly drawn from the testing datasets for each question. Example questions are available in the Appendix~\ref{app:user_study}.
% Figs.~\ref{fig:example_survey1} and Fig.~\ref{fig:example_survey2} (see Appendix~\ref{app:apendix}). 

% \paragraph{Confidence in Model Predictions} 
The first user study aimed to evaluate user overconfidence when assessing model predictions. Participants were shown an image along with the model's explanation and were asked to choose one of four response about the model's prediction. Answers included information wether the model's output was either correct or incorrect along with associated confidence level—categorized as fairly confident or somewhat confident. Results from this study are reported in Tab.~\ref{tab:user_study1}. The table reports key metrics on user's performance including true correct accuracy (user agreement when the model was right), true incorrect accuracy (user disagreement when the model was wrong), standard deviation and p-values assessing statistical significance compared to random guessing. The findings from this study reveal that participants exposed by explanations by \our{} exhibited general statistically significant confidence in the model’s correct predictions across ImageNet and CUB200-2011 datasets. However, users encounter challenges in accurately identifying incorrect predictions made by the model based on these explanations, a pattern consistent with previous findings from other XAI techniques.

\begin{table}[tbh]
\begin{minipage}[t]{0.48\textwidth}
    \centering
    \small
    \caption{The table reports metrics on the user's performance in the first user study, including accuracy and standard deviation. Statistically significant values are highlighted in bold. 
    % Results for ProtoPNet and GradCAM are referenced from~\cite{kim2022hive}, as indicated by the asterisk symbol.
    }
    \label{tab:user_study1}
    \renewcommand{\arraystretch}{1.2}
    \begin{tabular}{@{}l@{\;}c@{\;}c@{\;}c@{}}
        \toprule
        \textbf{Method} & \textbf{Prediction} & \textbf{ImageNet} & \textbf{CUB-200-2011} \\
        \midrule
        \multirow{2}{*}{\our{}} & Correct &  \textbf{0.637\textpm0.480} & \textbf{0.611\textpm0.487} \\
                                   & Incorrect & 0.447\textpm 0.497 & 0.294\textpm0.456 \\
        \midrule
        \multirow{2}{*}{InfoDisent} & Correct & \textbf{0.602\textpm0.090} & \textbf{0.807\textpm0.133} \\
                                   & Incorrect & \textbf{0.553\textpm0.099} & \textbf{0.427\textpm0.117} \\
        % \midrule
        \multirow{2}{*}{ProtoPNet*} & Correct & NA & 0.732\textpm0.249 \\
                                    & Incorrect & NA & \textbf{0.464\textpm0.359} \\
        % \midrule
        \multirow{2}{*}{GradCAM*} & Correct & 0.708\textpm0.266 & 0.724\textpm0.215 \\
                                  & Incorrect &\textbf{ 0.448\textpm0.316 }& 0.328\textpm0.243 \\
        \bottomrule
    \end{tabular}
\end{minipage}
\quad
\begin{minipage}[t]{0.48\textwidth}
    \centering
    \small
    \caption{The table reports accuracy, standard deviation and p-values for user's performance in second user study. 
    % We cite results for ProtoPNet and ProtoConcepts(~\cite{ma2023looks}), PIPNet and LucidPPN(~\cite{pach2024lucidppn}), as indicated by the asterisk symbol. 
    The p-value column indicates the p-value of a test against random.
    }
    \label{tab:user_study2}
    \renewcommand{\arraystretch}{1.2}
    \begin{tabular}{@{}l@{\;}c@{\;}c@{\;}c@{}}
        \toprule
        \textbf{Method} & \textbf{Dataset} & \textbf{User Acc.} & \textbf{p-value} \\
        \midrule
        \multirow{2}{*}{\our{}} & ImageNet & 0.568\textpm0.495 & $8 \cdot 10^{-4} $\\
                                    & CUB & 0.55\textpm0.497 & $9 \cdot 10^{-3}$ \\
        \midrule
        \multirow{2}{*}{InfoDisent} & ImageNet & 0.593\textpm0.149 & $8 \cdot 10^{-6}$ \\
                                    & CUB & 0.647\textpm0.131 & $10^{-14}$ \\
        % \midrule
        ProtoPNet* & CUB  & 0.515\textpm0.052 & 0.288 \\
        ProtoConcepts* & CUB & 0.621\textpm0.054 & $3 \cdot 10^{-5}$ \\
        PIPNet* & CUB & 0.600\textpm0.181 & 0.002 \\
        LucidPPN* & CUB & 0.679\textpm0.169 & $2 \cdot 10^{-6}$ \\
        \bottomrule
    \end{tabular}

\end{minipage}
    
\end{table}

% \paragraph{Disambiguity of prototypical parts}

The objective of the second user study was to evaluate how effectively participants could distinguish between prototypical parts. During the study, participants were presented with an image classified by the model, along with two explanations representing the top two most activated classes. Their task was to identify which class the model had ultimately selected, using only the information provided in the explanations. The results, shown in Tab.~\ref{tab:user_study2}, indicate that participants achieve statistically significantly higher accuracy in identifying the correct class for both the ImageNet and CUB200-2011 datasets compared to random guessing. This demonstrates that \our{} enhances user understanding of model predictions beyond mere chance levels. Details about the user study can be found in the Appendix~\ref{app:user_study}.
\section{Conclusions}
\label{sec:conclusions}

In this work, we introduced EPIC, a novel framework that unifies the strengths of post-hoc and prototype-based explanation methods for image classification. Unlike traditional prototype models that require specialized architectures and training procedures, EPIC operates directly on pretrained networks without altering their structure or predictions. At the same time, it retains the intuitive, human-understandable explanations offered by prototype-based approaches. Our experiments across benchmark and large-scale datasets demonstrate that EPIC provides high-quality, interpretable insights into model behavior while maintaining the flexibility and applicability of post-hoc methods. EPIC is a step toward making AI systems more transparent and easier to understand without sacrificing flexibility.

{ \bf Limitations} \our{} can be used only for architectures with a classification head consisting of a pooling layer on top of the backbone and a single-layer classification head.

% We have limited explanation in the case of the ViT model without a pooling layer.

% It is not obvious how to generalize such a model to different architectures. 
% In particular, our approach is currently limited to convolutional networks and has not yet been extended to transformer-based architectures such as ViT [Dosovitskiy et al., 2020] or multimodal foundation models like CLIP [Radford et al., 2021]. Recent works on transformer explainability [Chefer et al., 2021; Abnar & Zuidema, 2020], as well as concept-based explanations for CLIP [Gonzalez et al., 2022], could inspire future extensions of EPIC beyond CNN backbones.

% \bibliography{icml_main}
%\bibliography{references}
\bibliographystyle{plain}

%%%%%%%%%%%%%%%%%%%%%%%%%%%%%%%%%%%%%%%%%%%%%%%%%%%%%%%%%%%%

\appendix

\section{Appendix / supplemental material}
\label{app:apendix}

\subsection{Explanations of model decision}
\begin{figure}[tbh]
    \centering
    \qquad\qquad\qquad \our{} (our) \qquad\qquad\qquad\qquad\quad Grad-CAM \qquad\quad LRP \\
    \includegraphics[width=0.85\linewidth]{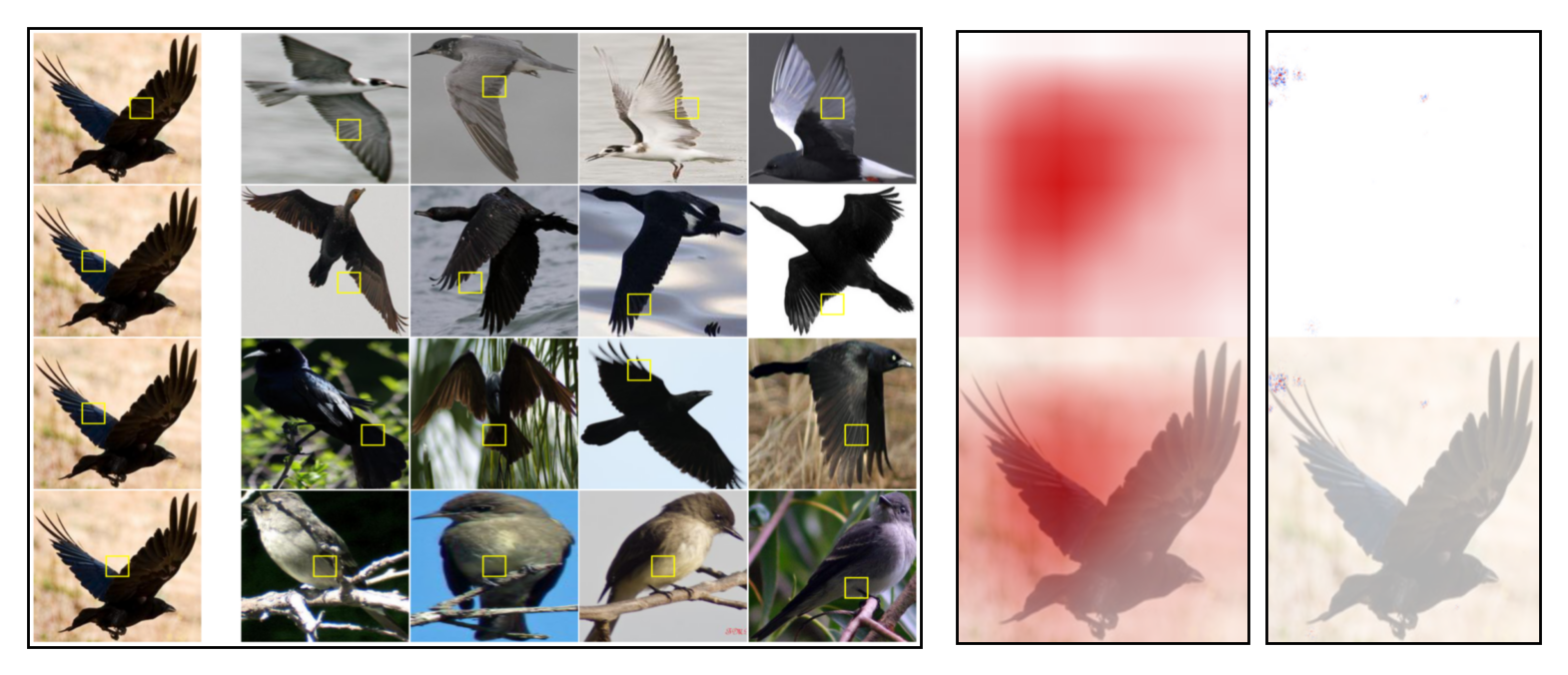}
    \includegraphics[width=0.85\linewidth]{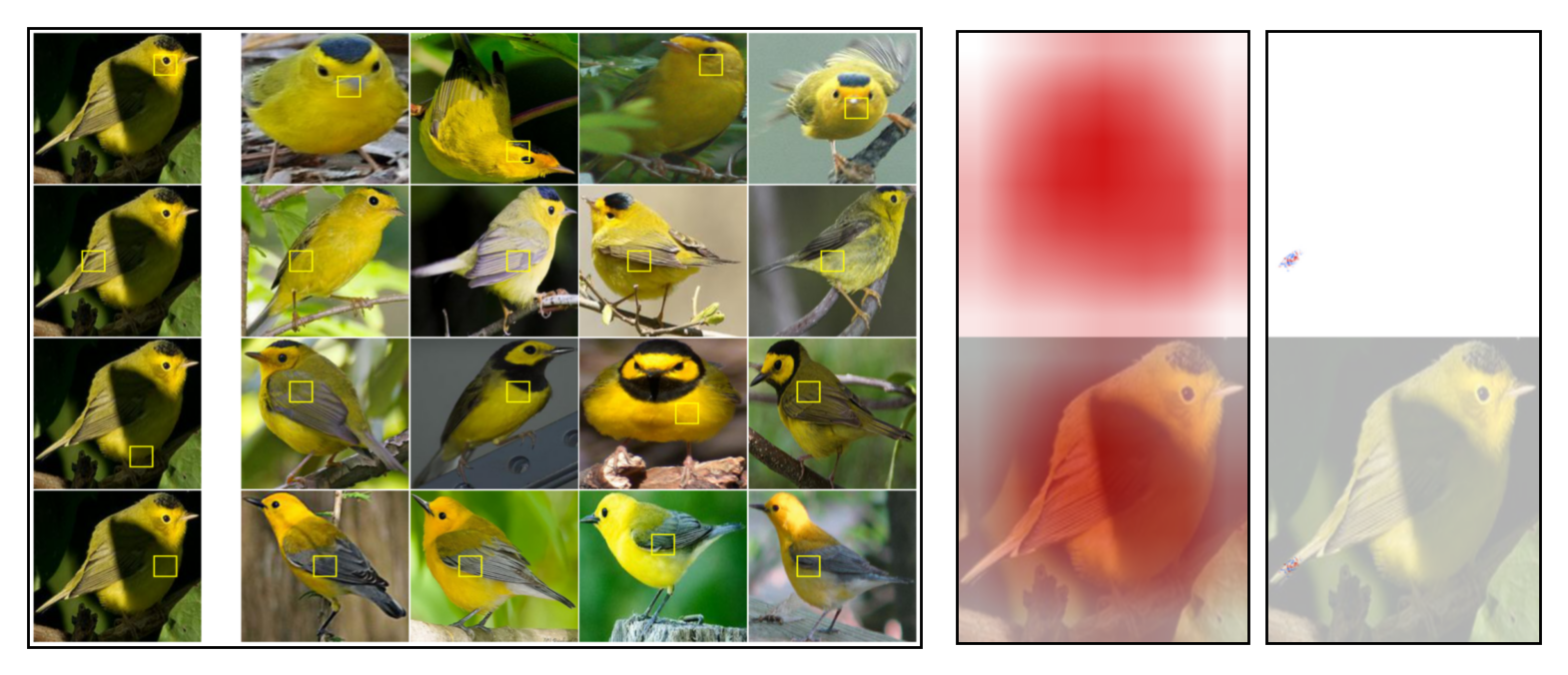}
    \includegraphics[width=0.85\linewidth]{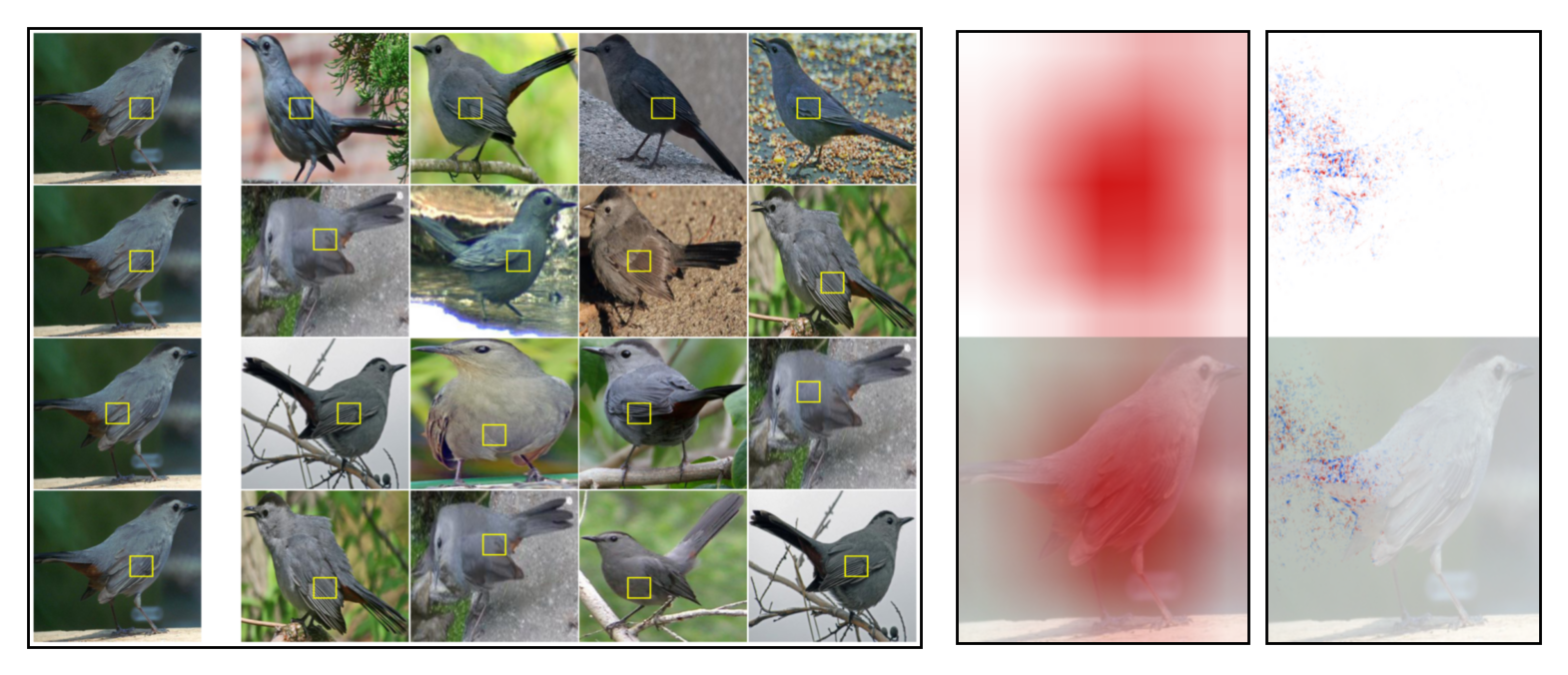}
    \includegraphics[width=0.85\linewidth]{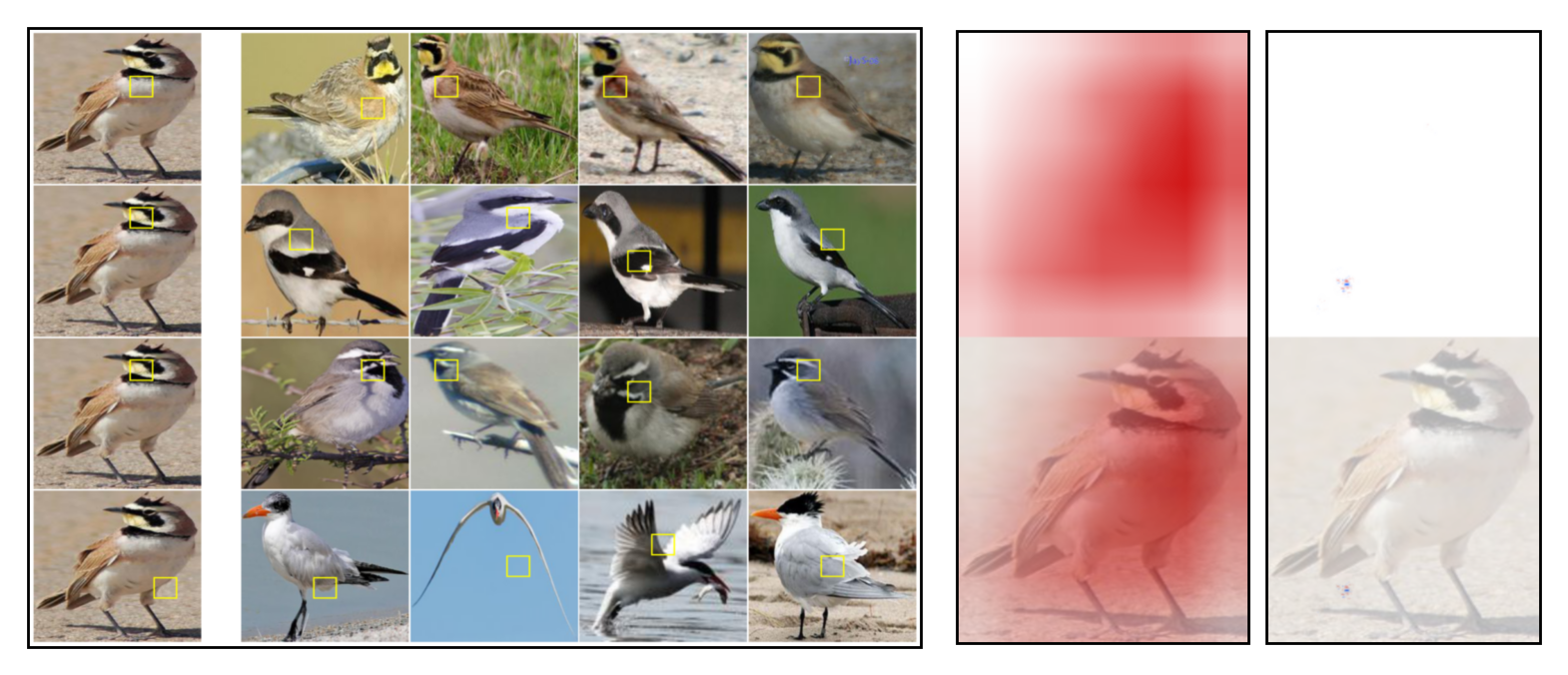}
    \caption{Comparison of explanations constructed by \our{}, and classical post-hoc models: Grad-CAM and LRP. The experiments were presented in the ResNet feature space on the images from the CUB200-2011 dataset.
    Each row represents the prototypical part. The yellow boxes in each row show the activation of a given prototypical part, while in the second column, we show the activation of corresponding prototypical parts in the original image.}
    \label{fig:app_explanations_cub}
\end{figure}

\begin{figure}[tbh]
    \centering
    \qquad\qquad\qquad \our{} (our) \qquad\qquad\qquad\qquad\quad Grad-CAM \qquad\quad LRP \\
    \includegraphics[width=0.85\linewidth]{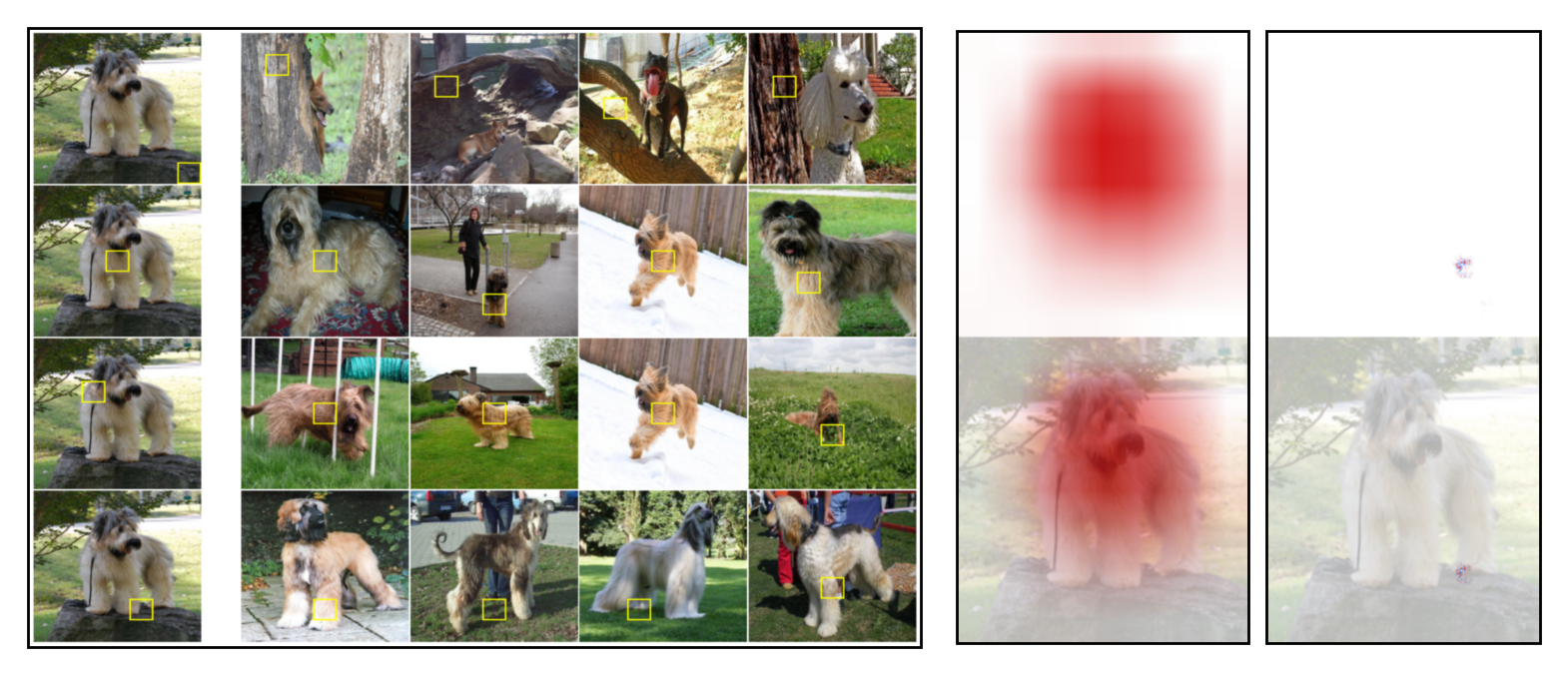}
    \includegraphics[width=0.85\linewidth]{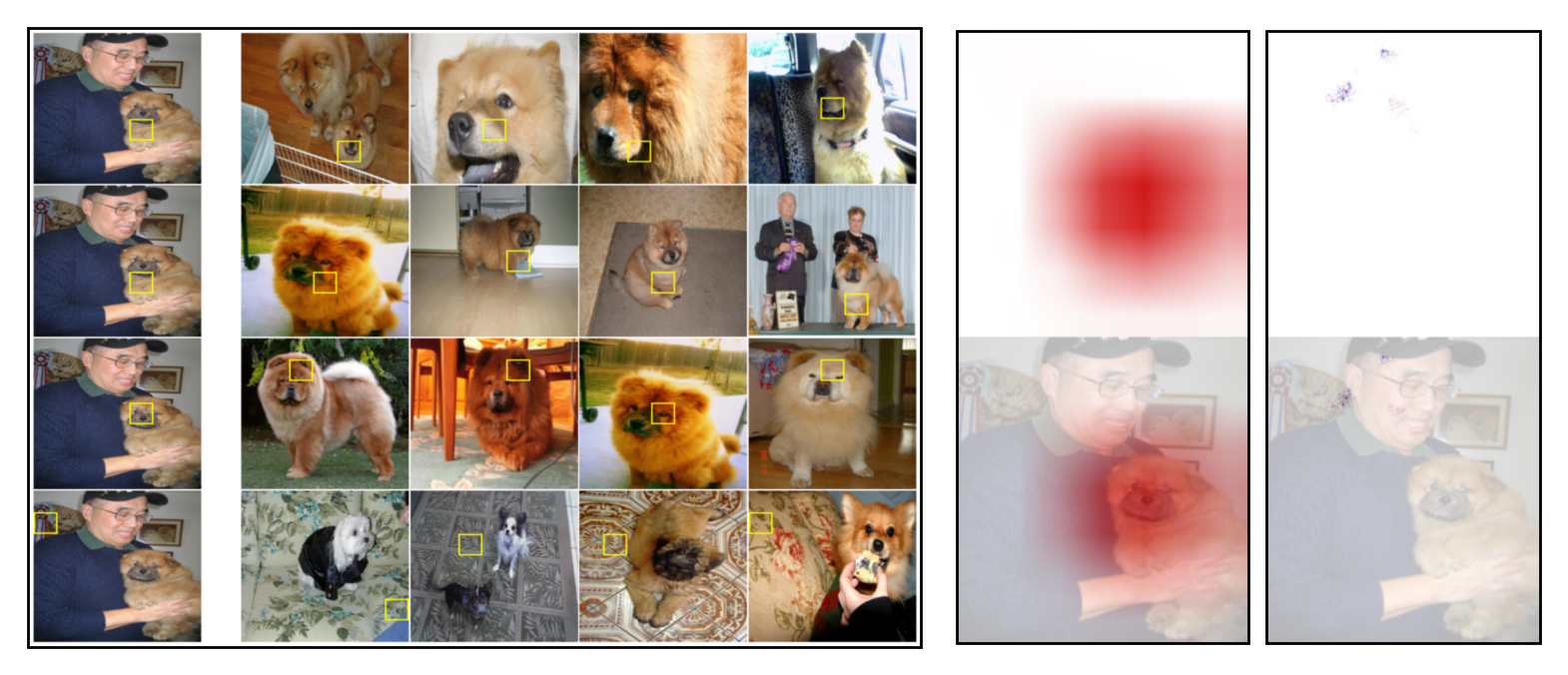}
    \includegraphics[width=0.85\linewidth]{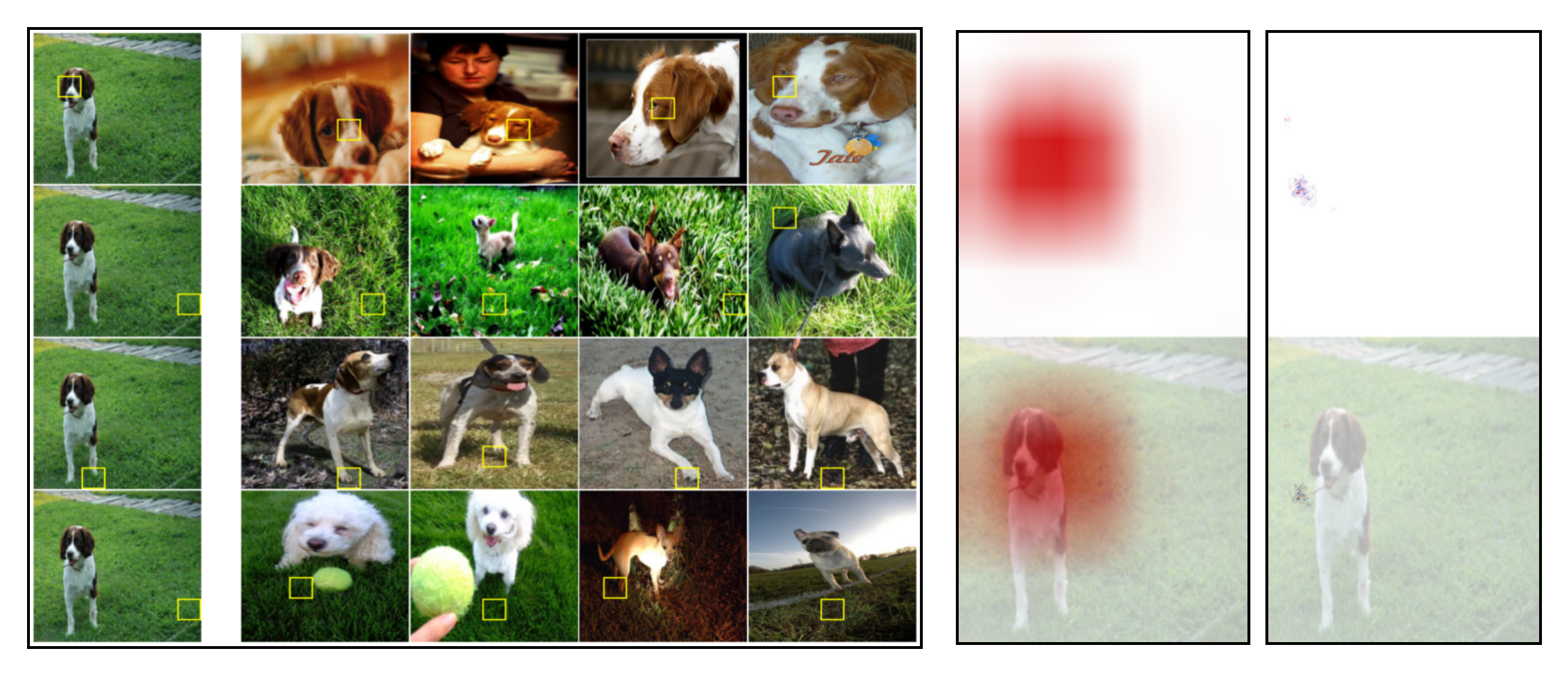}
    \includegraphics[width=0.85\linewidth]{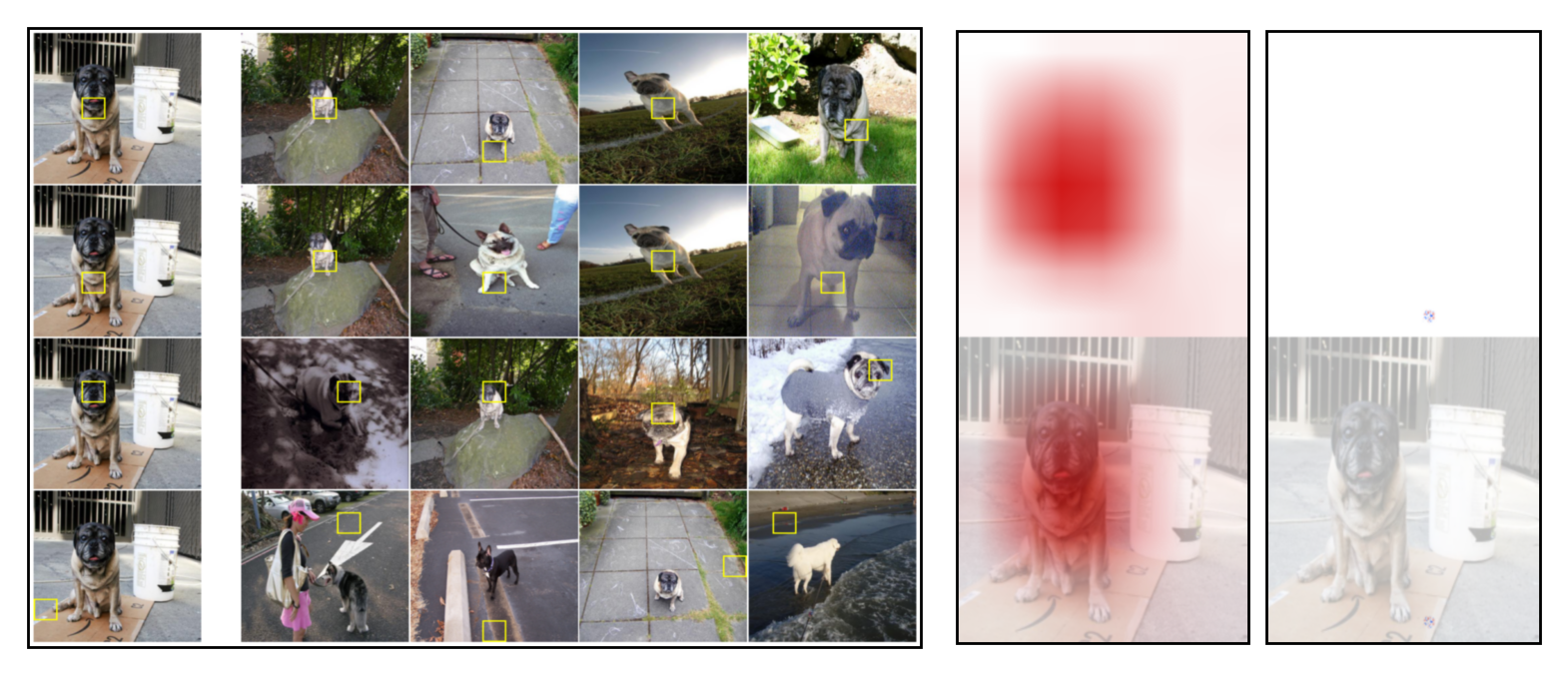}
    \caption{Comparison of explanations constructed by \our{}, and classical post-hoc models: Grad-CAM and LRP. The experiments were presented in the ResNet feature space on the images from the Stanford Dogs.
    Each row represents the prototypical part. The yellow boxes in each row show the activation of a given prototypical part, while in the second column, we show the activation of corresponding prototypical parts in the original image.}
    \label{fig:app_explanations_dogs}
\end{figure}

\begin{figure}[tbh]
    \centering
    \qquad\qquad\qquad \our{} (our) \qquad\qquad\qquad\qquad\quad Grad-CAM \qquad\quad LRP \\
    \includegraphics[width=0.85\linewidth]{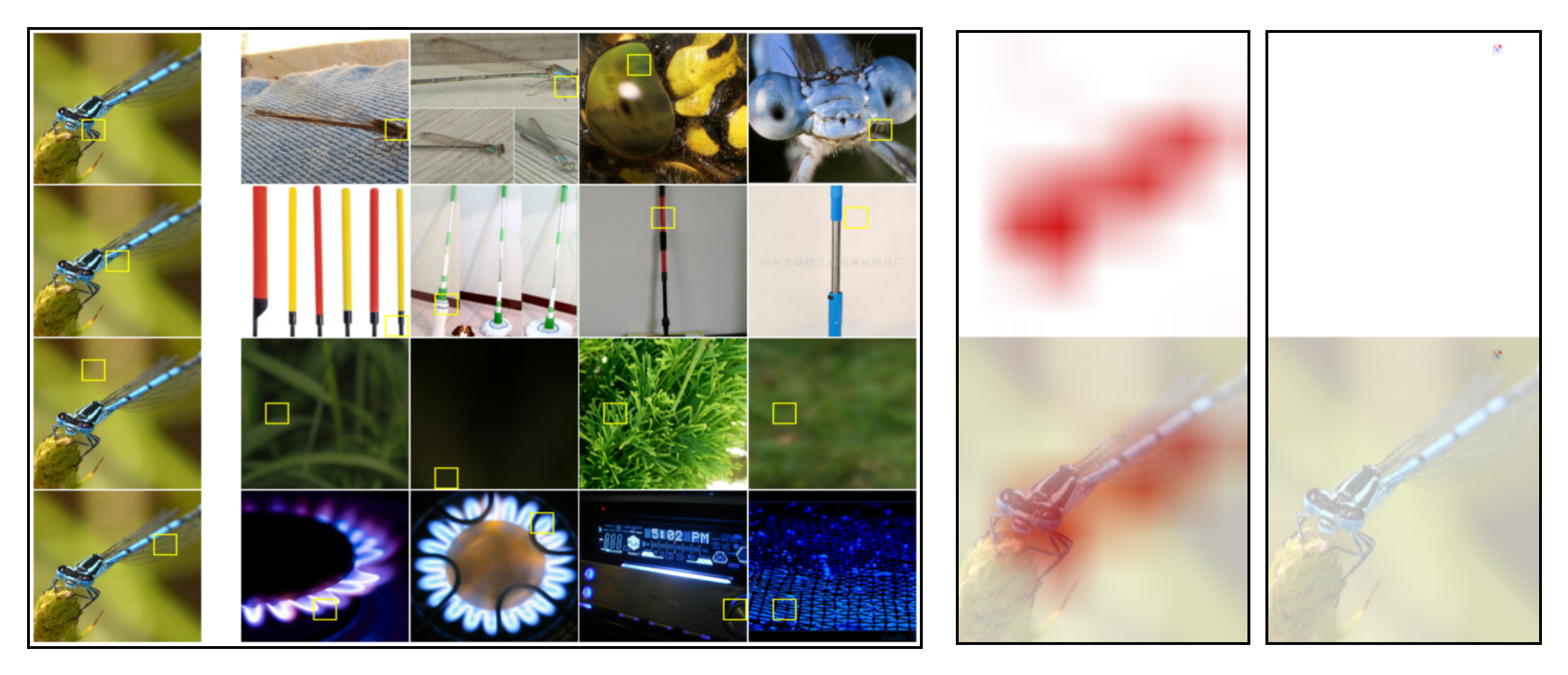}
    \includegraphics[width=0.85\linewidth]{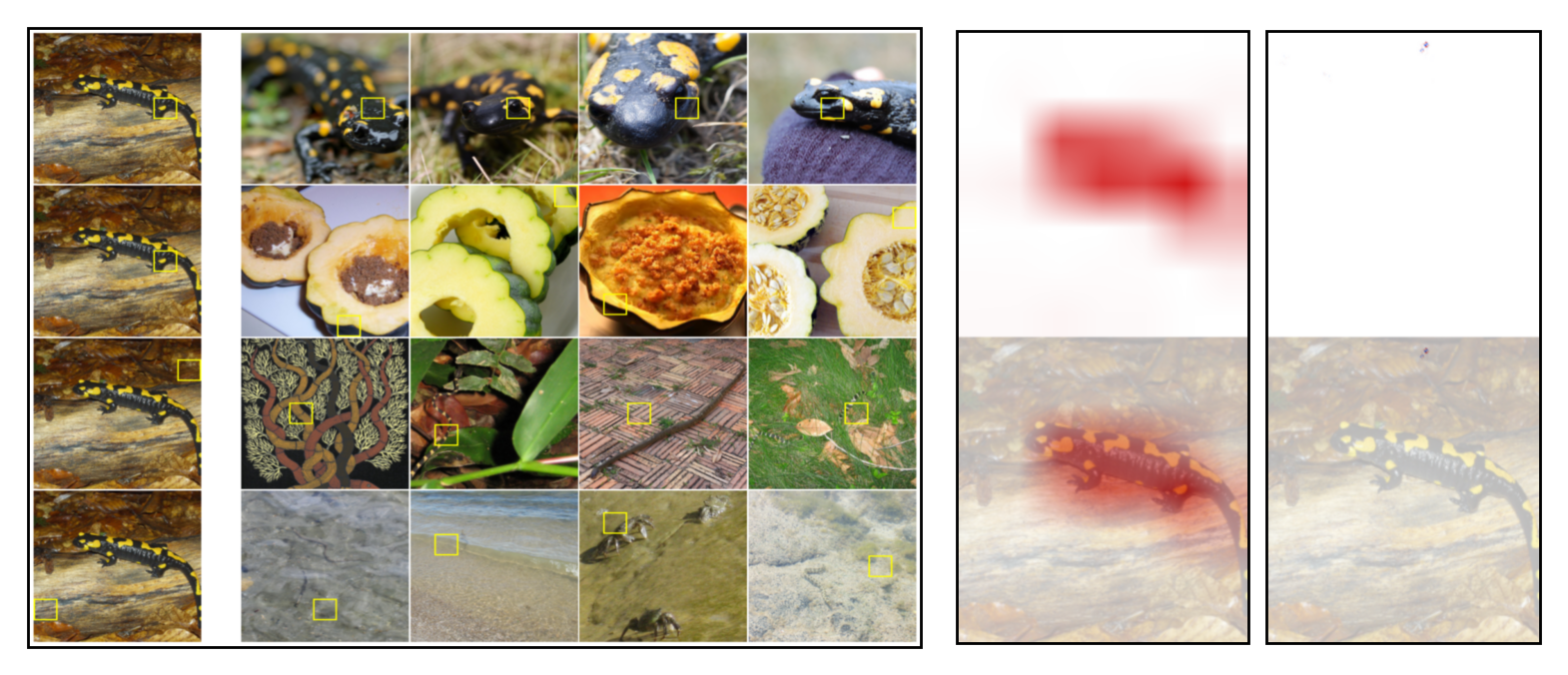}
    \includegraphics[width=0.85\linewidth]{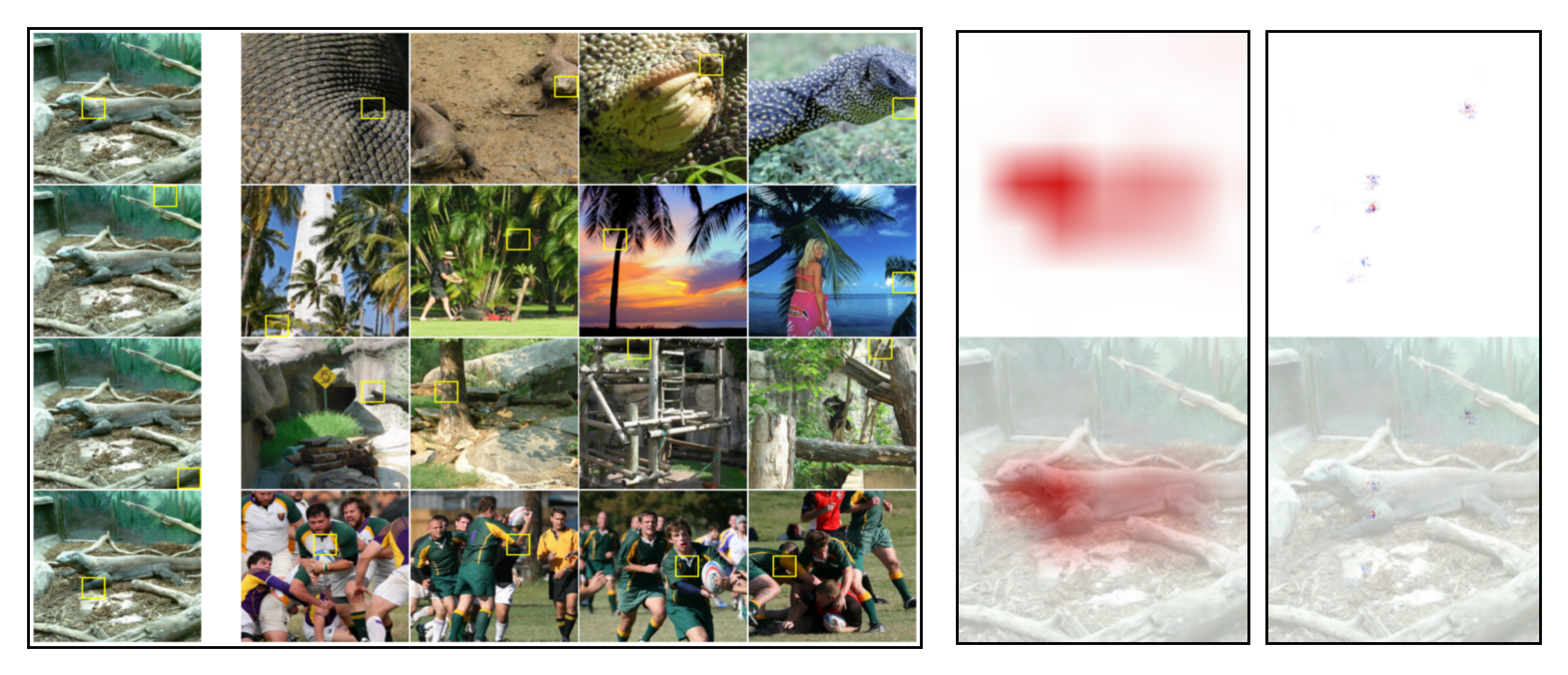}
    \includegraphics[width=0.85\linewidth]{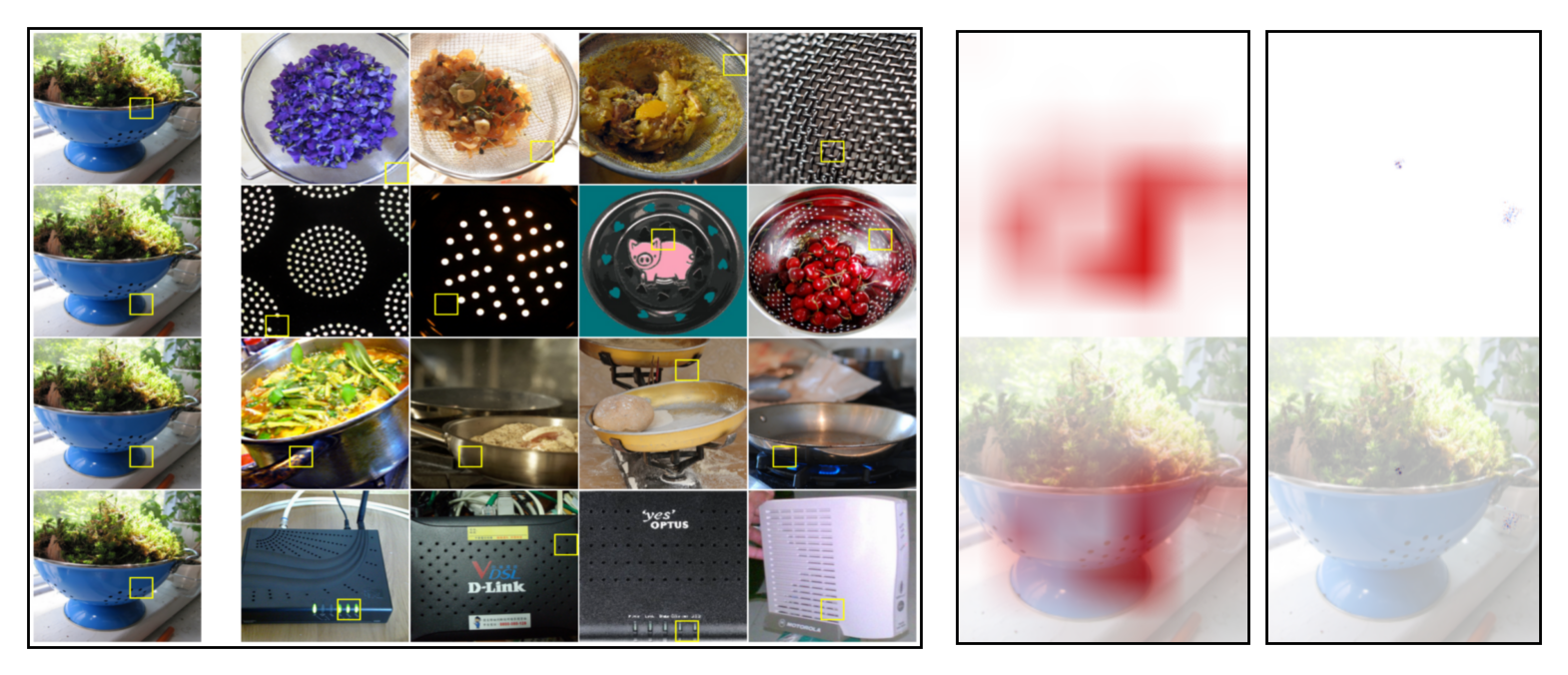}
    \caption{Comparison of explanations constructed by \our{}, and classical post-hoc models: Grad-CAM and LRP. The experiments were presented in the ResNet feature space on the images from the ImageNet dataset. Each row represents the prototypical part. The yellow boxes in each row show the activation of a given prototypical part, while in the second column, we show the activation of corresponding prototypical parts in the original image.}
    \label{fig:app_explanations_imagenet}
\end{figure}

In this section we provide additional results of experiments in explanations of model decision made by \our{} with its comparison to post-hoc approaches:  Grad-CAM and LRP. 
The experimental results are presented on the images from the CUB200-2011 (Fig.~\ref{fig:app_explanations_cub}), Stanford Dogs (Fig.~\ref{fig:app_explanations_dogs}) and ImageNet (Fig.~\ref{fig:app_explanations_imagenet}) datasets.

\subsection{More details on user study} 
\label{app:user_study}
\begin{figure}[thb]
    \centering
    \includegraphics[width=0.7\linewidth]{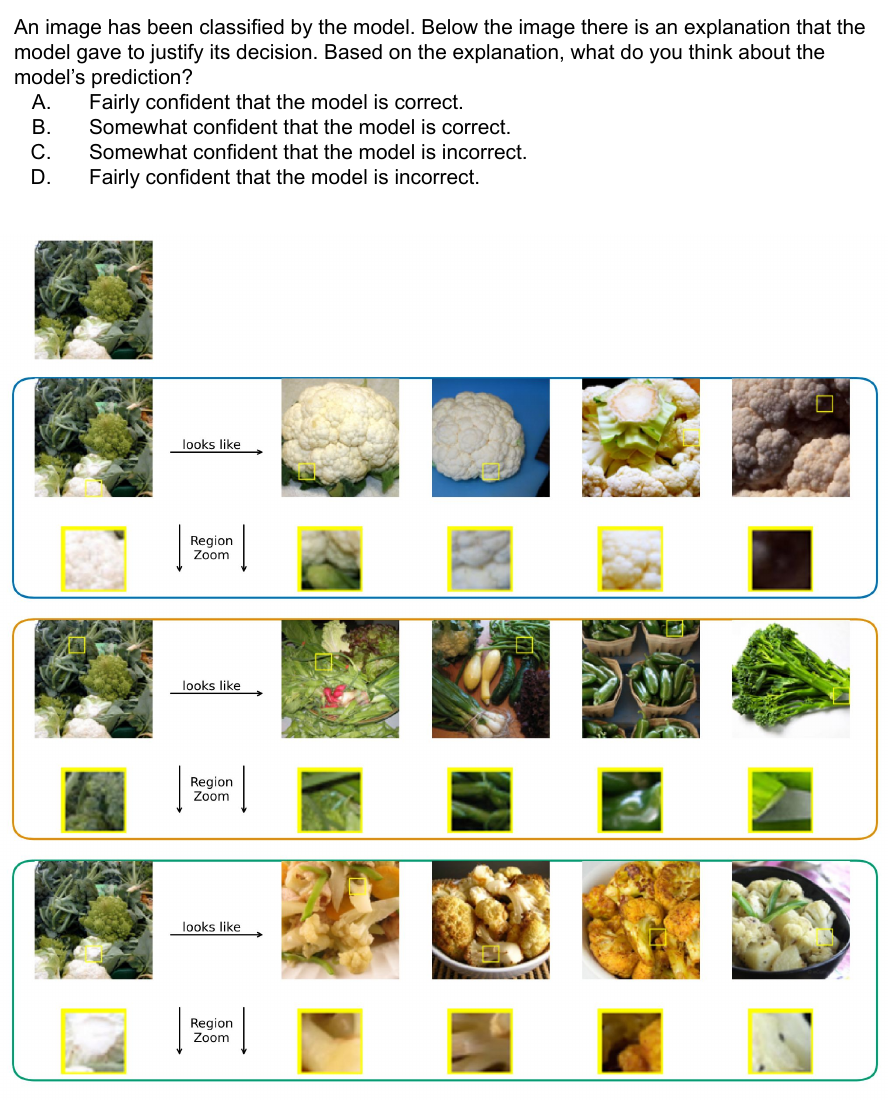}
    \caption{An exemplary question from the user study on user confidence.}
    \label{fig:example_survey1}
\end{figure}

\begin{figure}[thb]
    \centering
    \includegraphics[width=0.9\linewidth]{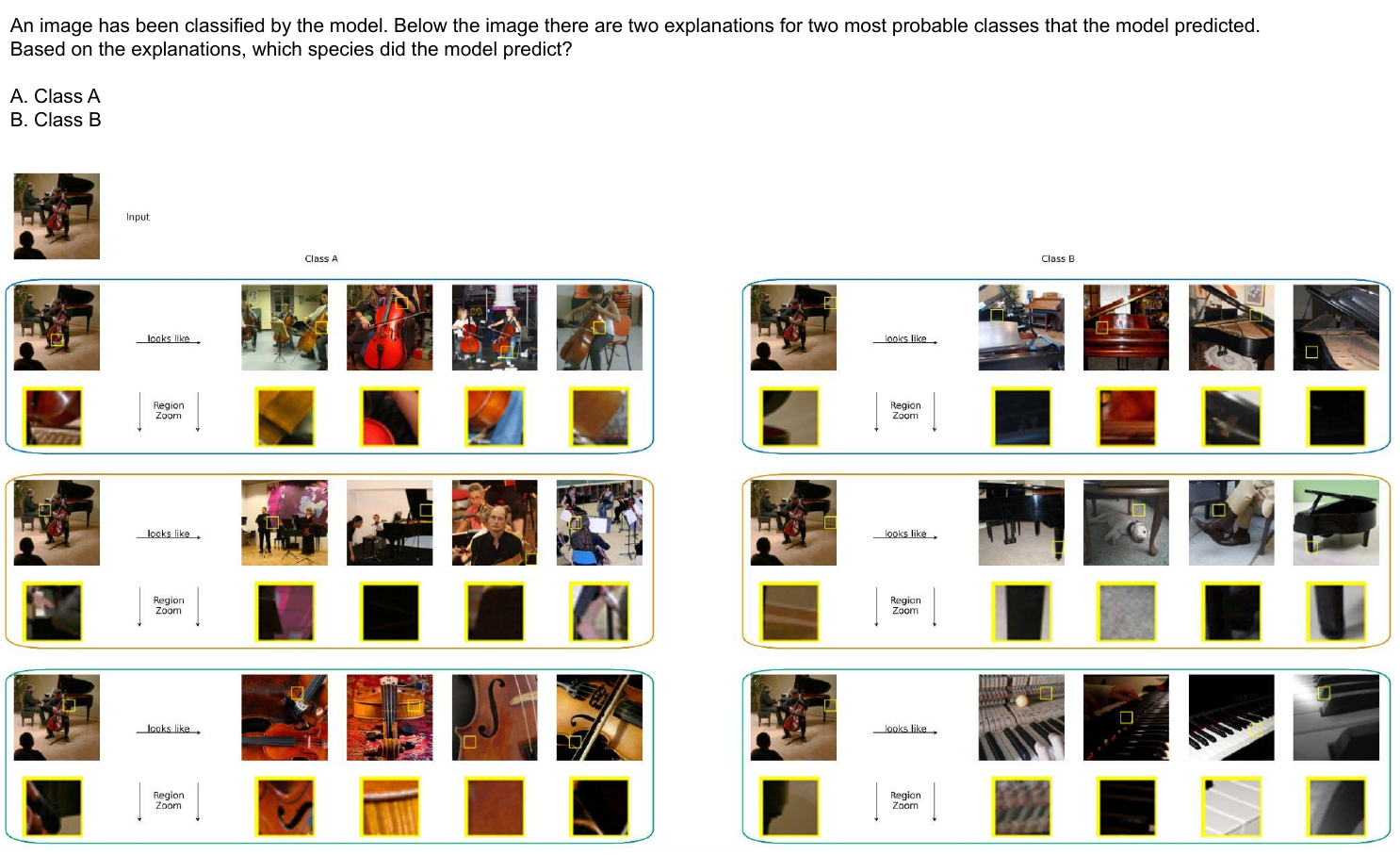}
    \caption{An exemplary question from the user study on disambiguity of prototypical parts.}
    \label{fig:example_survey2}
\end{figure}

In our usy studies the participants ranged in age from 18 to 60, with an average age of 35. Both studies were carried out on the Clickworker platform. Each worker was paid 2\euro{} for completing a short 20-question survey. The survey questions
were randomly composed, so the specific questions differed between participants. The participants were gender-balanced and
ranged in age from 18 to 60. They were given 30 minutes to complete the survey. To ensure data quality, we excluded responses where users selected the same answer for all questions. Surveys were repeated
until we obtained 60 valid responses for each dataset. Fig.~\ref{fig:example_survey1} and Fig.~\ref{fig:example_survey2}  illustrate example questions used in both user studies.
Before starting the survey, participants were provided with an example and detailed instructions to familiarize them with the
study setup, including the explanation composition and visualization. The distribution of answers is summarized in Tab.~\ref{tab:user_study1}
and Tab.~\ref{tab:user_study2}.

\subsection{Classification Performance}
\label{app:class_perform}
As previously discussed, the design of \our{} maintains the predictive performance of the pretrained model. In other words, integrating the \module{} yields the same output for an image $I$ as the original model. While additional operations could potentially introduce numerical errors, we demonstrate that this is not the case by reporting numerical accuracy on CUB-200-2011, Stanford Dogs, Stanford Cars, see Tab.~\ref{tab:acc_split_tables} and Tab.~\ref{tab:acc_cropped_data}.

\subsection{Datasets}

In our experiments we utilized four key datasets: ImageNet~\cite{5206848}, Stanford Cars~\cite{krause2013collecting}, Stanford Dogs~\cite{KhoslaYaoJayadevaprakashFeiFei_FGVC2011}, CUB200-2011~\cite{WahCUB_200_2011}, which are frequently employed in prototype model evaluations. All of these datasets contain large-scale image collections and fine-grained class distinctions. The datasets' high intra-class similarities pose significant challenges for prototype-based models. It is worth noting that only one of the previous prototypical parts-based methods, namely InfoDiscent~\cite{struski2024infodisent}, has been generalized to the ImageNet dataset. Comparison between \our{} and InfoDiscent is presented in Fig.~\ref{fig:comp_2}.

\begin{table*}[t]
  \centering
  \caption{Classification accuracy on full CUB-200-2011, and Stanford Dogs datasets by competing approaches using different CNN backbones. For each dataset and backbone, we boldface the best result in the class of interpretable models.}
  \begin{minipage}[t]{0.32\textwidth}
    \centering
    \textbf{ResNet-34}\\[0.5ex]
    \begin{tabular}{@{}l@{\,}c@{\,}c@{}}
      \toprule
      Model & CUB & Dogs \\
      \midrule
      ResNet-34 & 76.0\% & 84.5\% \\
      \our{\,(ours)} & 76.0\% & \textbf{84.5}\% \\
      InfoDisent & \textbf{78.3}\% & 83.9\% \\
      \cdashlinelr{1-3}
      ProtoPNet & 74.1\% & 76.1\% \\
      ST-ProtoPNet & 78.2\% & 83.4\% \\
      TesNet & 76.5\% & 81.2\% \\
      \bottomrule
    \end{tabular}
  \end{minipage}
  \hfill
  \begin{minipage}[t]{0.32\textwidth}
    \centering
    \textbf{ResNet-50}\\[0.5ex]
    \begin{tabular}{@{}l@{\,}c@{\,}c@{}}
      \toprule
      Model & CUB & Dogs \\
      \midrule
      ResNet-50 & 78.7\% & 87.4\% \\
      \our{\,(ours)} & 78.7\% & \textbf{87.4}\% \\
      InfoDisent & 79.5\% & 86.6\% \\
      \cdashlinelr{1-3}
      ProtoPNet & 84.8\% & 78.1\% \\
      ST-ProtoPNet & \textbf{88.0}\% & 83.3\% \\
      TesNet & 87.3\% & 85.7\% \\
      \bottomrule
    \end{tabular}
  \end{minipage}
  \hfill
  \begin{minipage}[t]{0.32\textwidth}
    \centering
    \textbf{DenseNet-121}\\[0.5ex]
    \begin{tabular}{@{}l@{\,}c@{\,}c@{}}
      \toprule
      Model & CUB & Dogs \\
      \midrule
      DenseNet-121 & 78.2\% & 84.1\% \\
      \our{\,(ours)} & 78.2\% & 75.4\% \\
      InfoDisent & 80.6\% & \textbf{83.8}\% \\
      \cdashlinelr{1-3}
      ProtoPNet & 76.6\% & 75.4\% \\
      ST-ProtoPNet & \textbf{81.8}\% & 82.9\% \\
      TesNet & 80.9\% & 82.1\% \\
      \bottomrule
    \end{tabular}
  \end{minipage}
  \label{tab:acc_split_tables}
\end{table*}

\subsection{Experiments details}
All experiments were conducted on an NVIDIA GeForce RTX 4090 GPU. The process of finding a set of prototypes is highly dependent on the size of the training set. For ImageNet, training takes up to 16 hours on a single GPU on the larger models.

\subsection{Explaining model prediction}
~\label{app:model_pred}
\begin{figure}[tbh]
    \centering
    \qquad \our{} (our) \qquad\qquad\qquad\qquad\qquad\qquad InfoDisent \\
    \includegraphics[width=0.45\linewidth]{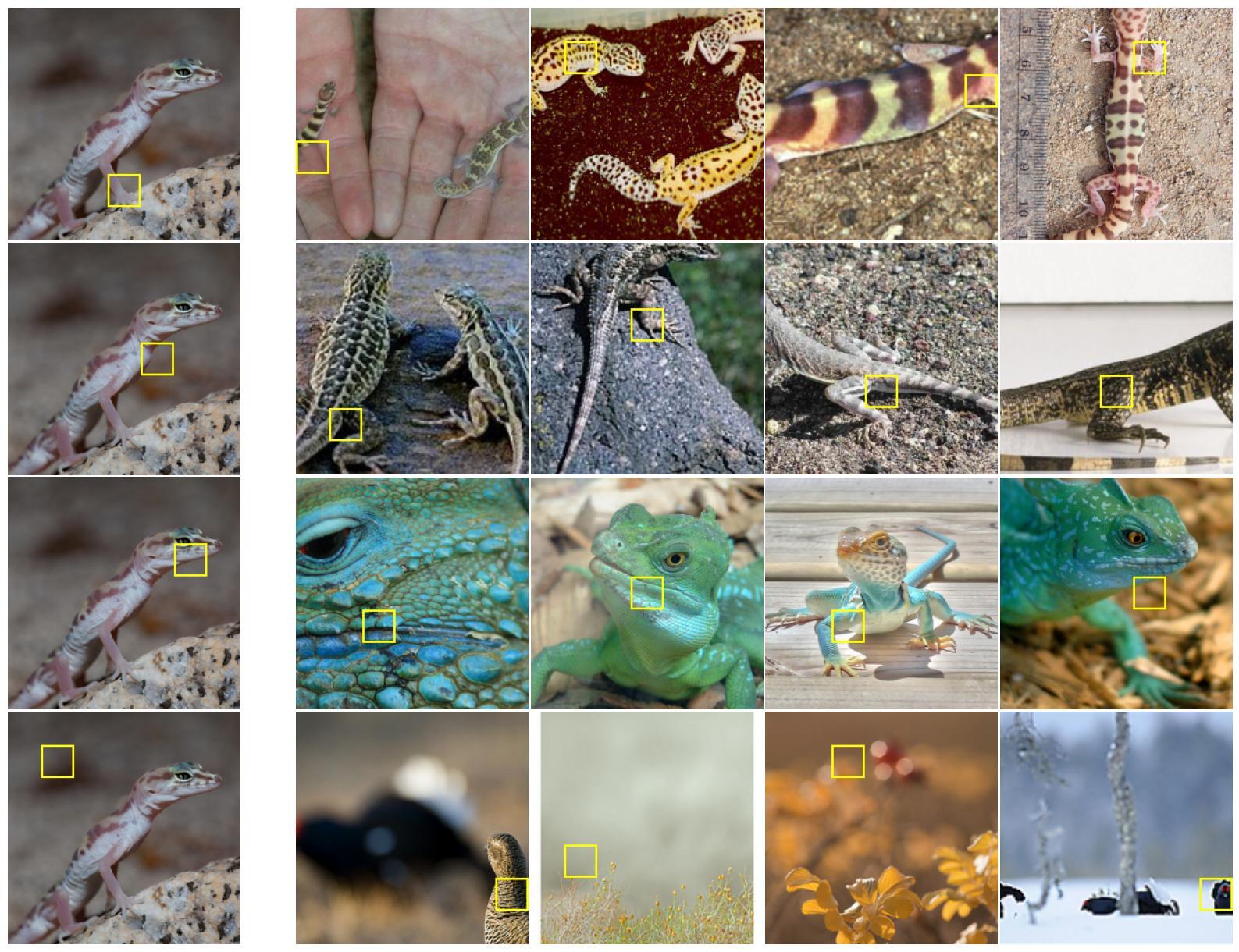}
        \quad
        \includegraphics[width=0.45\linewidth]{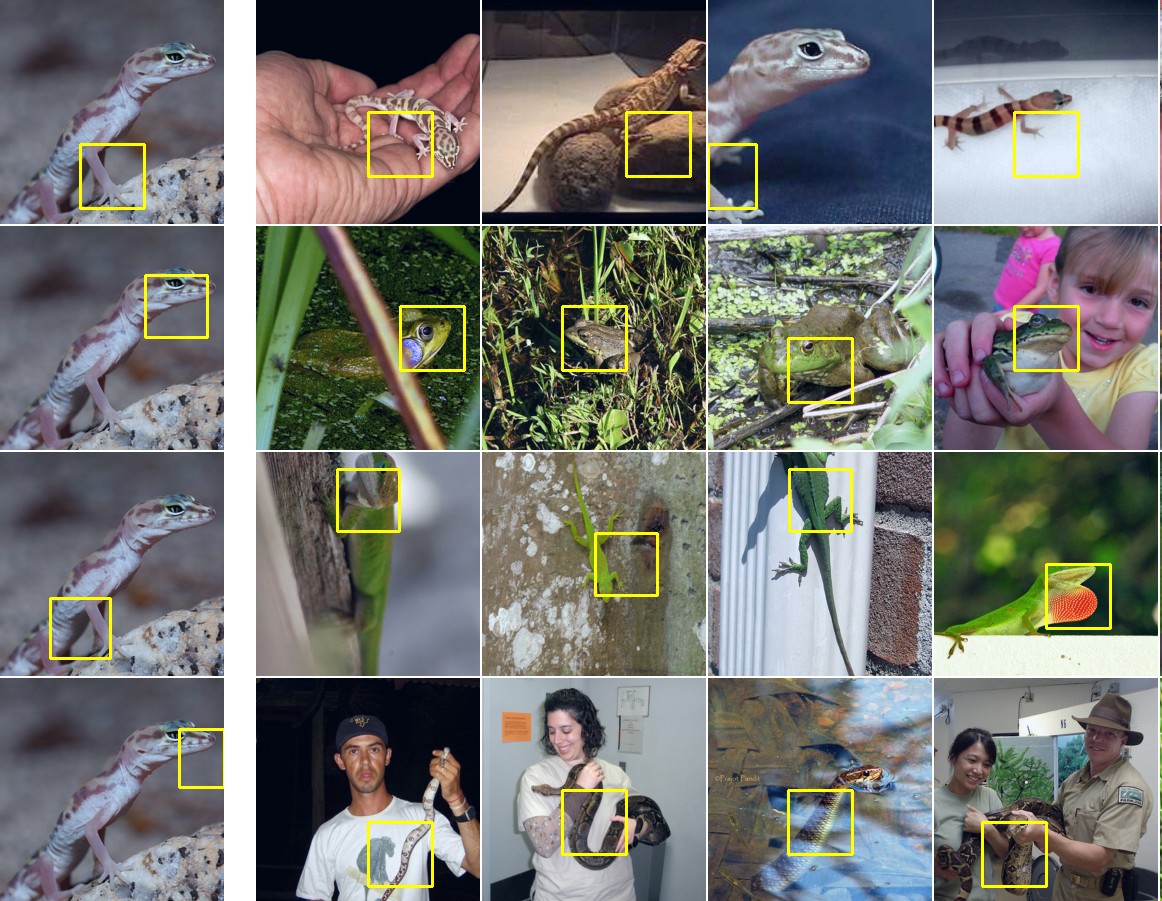}
    \\
        \includegraphics[width=0.45\linewidth]{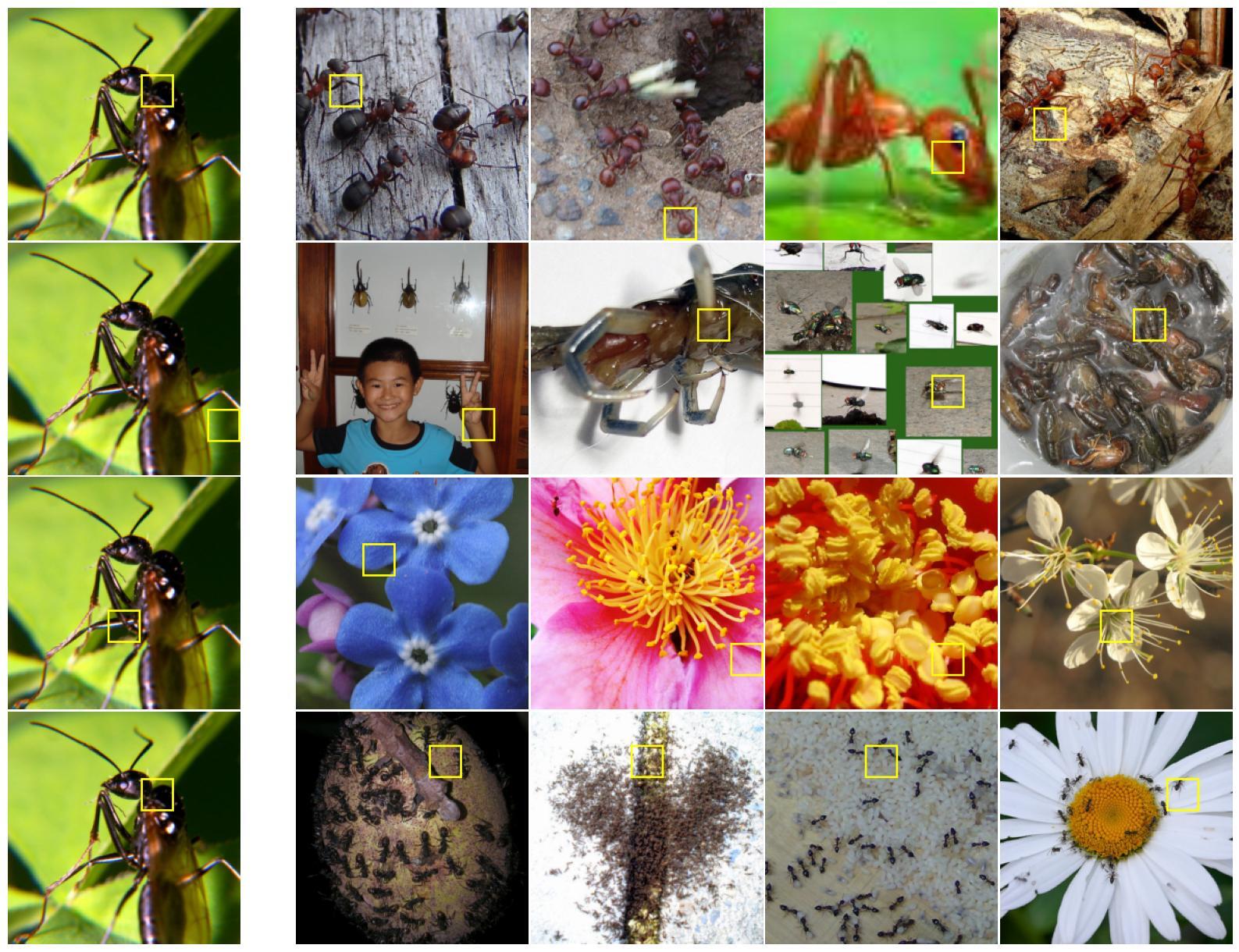}
        \quad
        \includegraphics[width=0.45\linewidth]{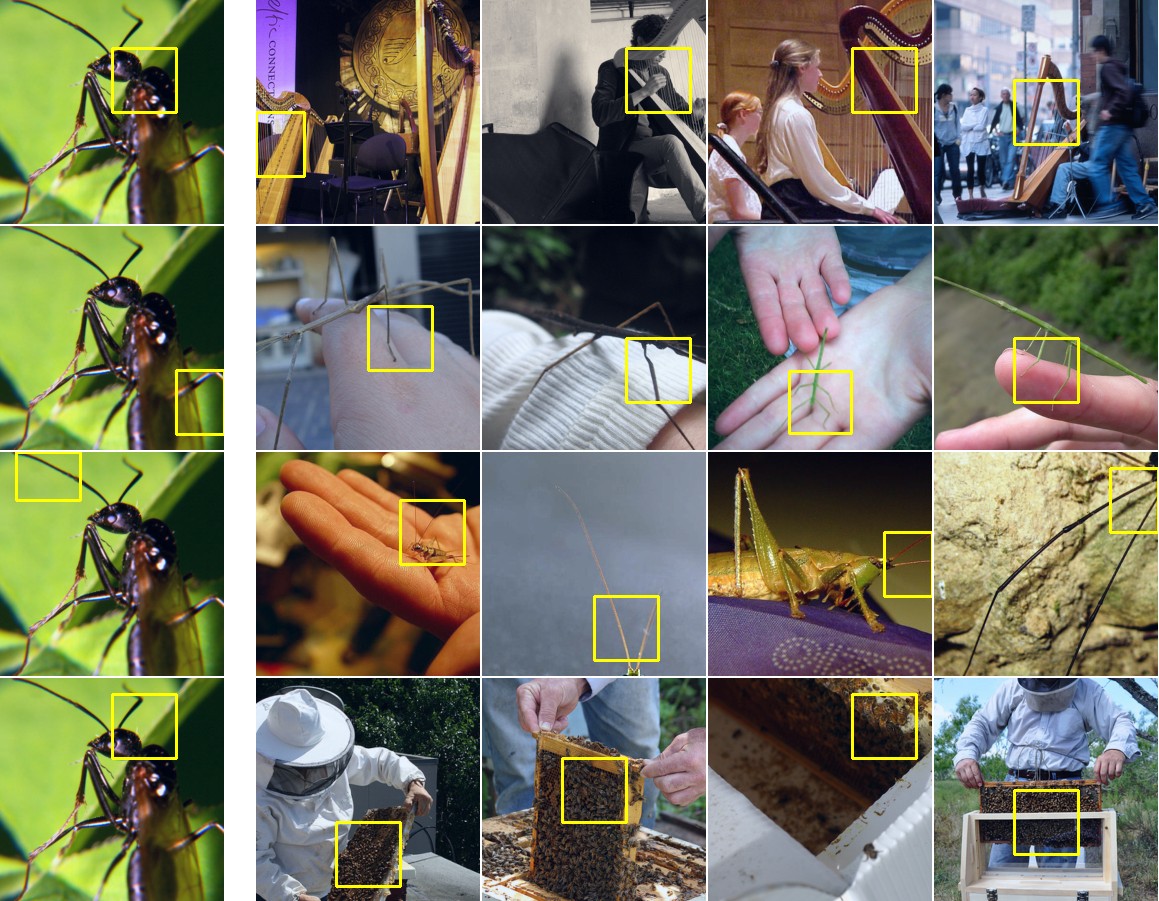}
    \\
%     \caption{Comparison of explanations between \our{} (our) and prototype-based model InfoDisent. InfoDisent works on top of the pretrain backbone and can give predictions for the ImageNet dataset. \our{} build prototypes more connected with input images.   }
%     \label{fig:comp_ant}
% \end{figure}

% \begin{figure}[!t]
%     \centering
%     \qquad \our{} (our) \qquad\qquad\qquad\qquad\qquad\qquad InfoDisent
    % \\
        \includegraphics[width=0.45\linewidth]{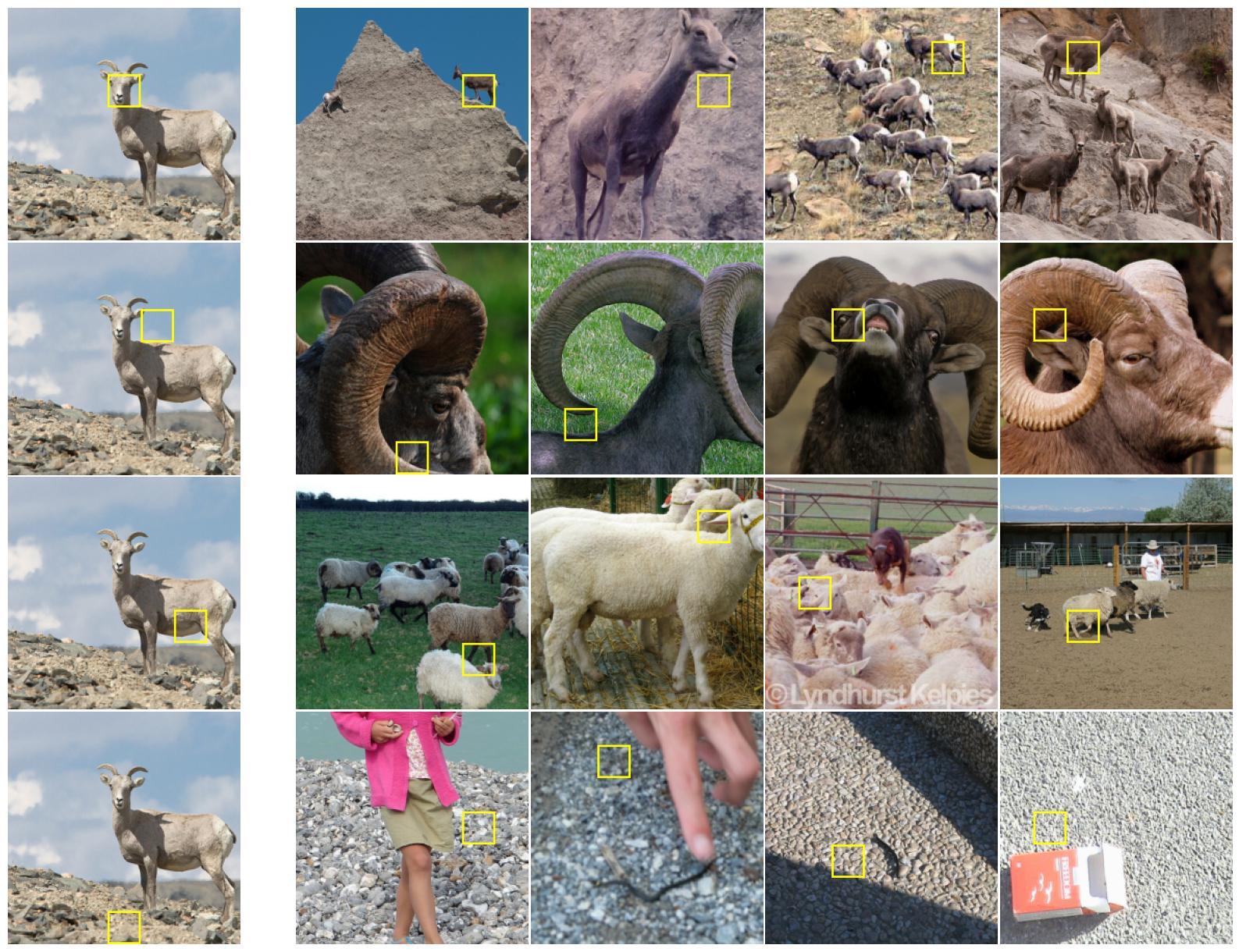}
        \quad
        \includegraphics[width=0.45\linewidth]{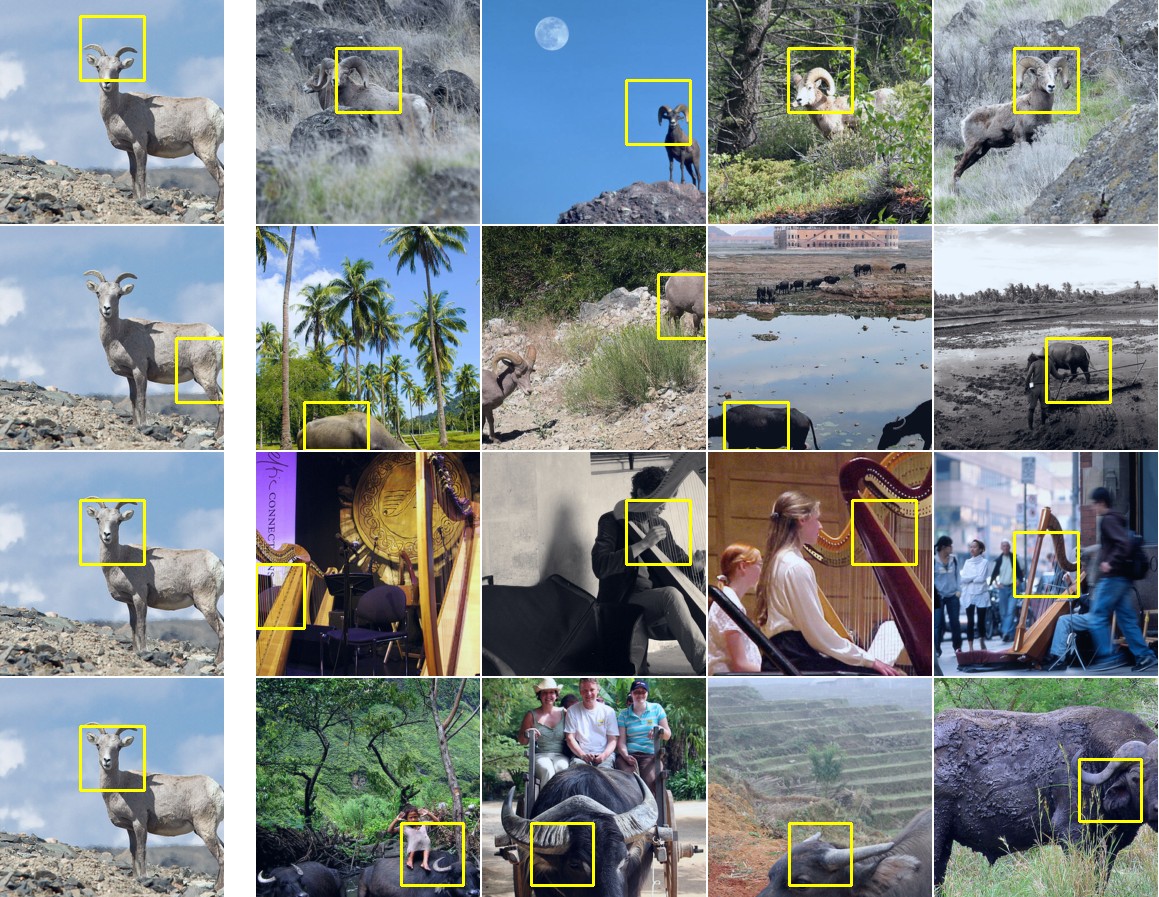}
    \\
        \includegraphics[width=0.45\linewidth]{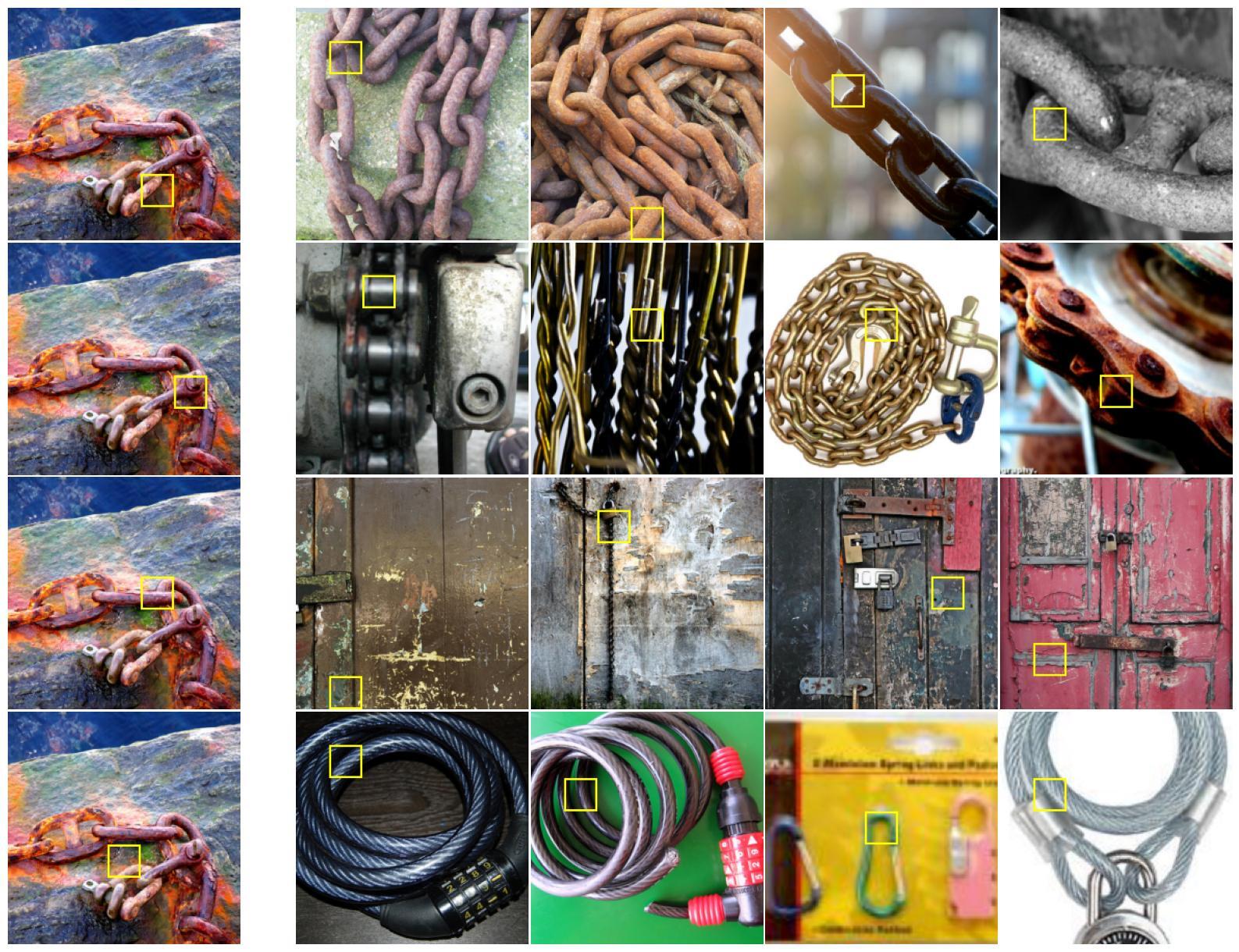}
        \quad
        \includegraphics[width=0.45\linewidth]{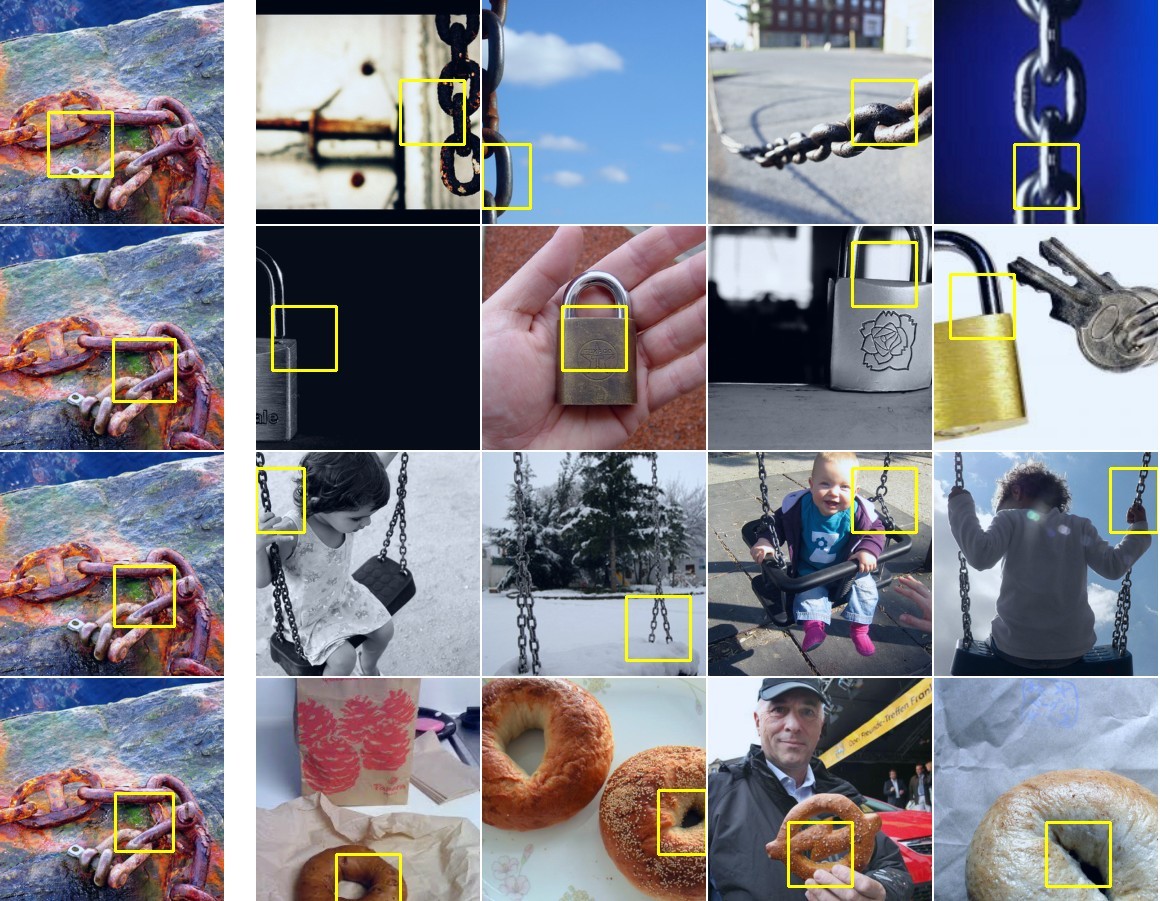}
    \caption{Comparison of explanations between \our{} (our) and prototype-based model InfoDisent. InfoDisent works on top of the pretrain backbone and can give predictions for the ImageNet dataset. \our{} build prototypes more connected with input images.   }
    \label{fig:comp_2}
\end{figure}

After completing the training of the \module{} and selecting the channel prototypes, the next step is to explain the model's prediction for a given input image. This is achieved by selecting $k$ channels with the highest contribution to the predicted class. This can be done by examining the terms contributing to the model output in the final classification layer. More precisely, for an input image $I$ and the model prediction of the input belonging to class $y$, we follow the algorithm outlined in Algorithm \ref{alg:explain}. Since we are only interested in the positive prototypes, we apply $\relu$ before examining the terms contributing to the sum. Example explanation is shown in Fig. \ref{fig:comp_2}.

\begin{table*}[t]
  % \small
  \centering
  \caption{Accuracy comparison of interpretability models using standard CNN architectures (utilized in explainable models) trained on cropped bird images of CUB-200-2011, and Stanford Cars (Cars). Our approach demonstrates superior performance across nearly all the datasets and models considered. For each dataset and backbone, we boldface the best result in the class of interpretable models.}
  \begin{minipage}[t]{0.49\textwidth}
    \centering
    \textbf{ResNet-34}\\[0.45ex]
    \begin{tabular}{@{}l@{\,}c@{\,}c@{}}
      \toprule
      Model & CUB & Cars \\
      \midrule
      ResNet-34 & 82.4\% & 92.6\% \\
      \our{\,(ours)} & 82.4\% & 92.6\% \\
      InfoDisent & \textbf{83.5}\% & \textbf{92.8}\% \\
      \cdashlinelr{1-3}
      ProtoPNet & 79.2\% & 86.1\% \\
      ProtoPShare & 74.7\% & 86.4\% \\
      ProtoPool & 80.3\% & 89.3\% \\
      ST-ProtoPNet & \textbf{83.5}\% & 91.4\% \\
      TesNet & 82.7\% & 90.9\% \\
      \bottomrule
    \end{tabular}
  \end{minipage}
  % \hfill
    \begin{minipage}[t]{0.49\textwidth}
    \centering
    \textbf{DenseNet-121}\\[0.5ex]
    \begin{tabular}{@{}l@{\,}c@{\,}c@{}}
      \toprule
      Model & CUB & Cars \\
      \midrule
      DenseNet-121 & 81.8\% & 92.1\% \\
      \our{\,(ours)} & 81.8\% & 92.1\% \\
      InfoDisent & 82.6\% & \textbf{92.7}\% \\
      \cdashlinelr{1-3}
      ProtoPNet & 79.2\% & 86.8\% \\
      ProtoPShare & 74.7\% & 84.8\% \\
      ProtoPool & 73.6\% & 86.4\% \\
      ST-ProtoPNet & \textbf{85.4}\% & 92.3\% \\
      TesNet & 84.8\% & 92.0\% \\
      \bottomrule
    \end{tabular}
  \end{minipage} \\[2ex]
  % \hfill
  \begin{minipage}[t]{0.49\textwidth}
    \centering
    \textbf{ResNet-50}\\[0.45ex]
    \begin{tabular}{@{}l@{\,}c@{\,}c@{}}
      \toprule
      Model & CUB & Cars \\
      \midrule
      ResNet-50 & 83.2\% & 93.1\% \\
      \our{\,(ours)} & \textbf{83.2}\% & \textbf{93.1}\% \\
      InfoDisent & 83.0\% & 92.9\% \\
      \cdashlinelr{1-3}
      ProtoPool & -- & 88.9\% \\
      ProtoTree & -- & 86.6\% \\
      PIP-Net & 82.0\% & 86.5\% \\
      \bottomrule
    \end{tabular}
  \end{minipage}
  % \hfill
  \begin{minipage}[t]{0.49\textwidth}
    \centering
    \textbf{ConvNeXt}\\[0.5ex]
    \begin{tabular}{@{}l@{\,}c@{\,}c@{}}
      \toprule
      Model & CUB & Cars \\
      \midrule
      ConvNeXt-Tiny & 83.8\% & 91.0\% \\
      \our{\,(ours)} & 83.8\% & \textbf{91.0\%} \\
      InfoDisent & 84.1\% & 90.2\% \\
      \cdashlinelr{1-3}
      PIP-Net & \textbf{84.3}\% & 88.2\% \\
      \bottomrule
    \end{tabular}
  \end{minipage}
  \label{tab:acc_cropped_data}
\end{table*}

%%%%%%%%%%%%%%%%%%%%%%%%%%%

\begin{algorithm}
\caption{Top-$k$ Contributing Channels}
\label{alg:explain}
\begin{algorithmic}[1]
\Procedure{TopKContributingChannels}{$\Phi_\Theta$, $A$, $I$, $k$, $U$}
    \State $Z \gets \Phi_\Theta(I) \in \mathbb{R}^{H \times W \times D}$ \Comment{Feature map}
    \State $\hat{Z} \gets U \circledast Z \in \mathbb{R}^{H \times W \times D}$ \Comment{\module{}}
    \State $A' \gets A U^{-1}$

    \State $v \gets \avgpoolch(\hat{Z}) \in \mathbb{R}^D$ \Comment{Global average pooling}
    \State $w \gets A' v \in \mathbb{R}^C$ \Comment{Logits}
    \State $\text{pred} \gets \arg\max(w)$ \Comment{Predicted class}
    \State $w_{\text{pred}} \gets A'[\text{pred}]$ \Comment{Weights for predicted class}
    \State $\text{scores} \gets w_{\text{pred}} \circledast \operatorname{ReLU}(v)$ \Comment{Element-wise product}
    \State $\text{channels} \gets \text{TopK}(\text{scores}, k)$
    \State \textbf{return} $\text{channels}$
\EndProcedure
\end{algorithmic}
\end{algorithm}

%%%%%%%%%%%%%%%%%%%%%%%%%%%%%%%%%%%%%%%%%%%%%%%%%%%%%%

\end{document}